\documentclass{article}

\PassOptionsToPackage{numbers, compress}{natbib}
\usepackage[preprint]{neurips_2026}

\usepackage[utf8]{inputenc} 
\usepackage[T1]{fontenc}    
\usepackage{hyperref}       
\usepackage{url}            
\usepackage{booktabs}       
\usepackage{amsfonts}       
\usepackage{nicefrac}       
\usepackage{microtype}      
\usepackage{xcolor}         
\usepackage{float}  
\usepackage{microtype}
\usepackage{graphicx}
\usepackage{subcaption}
\usepackage{booktabs} 
\usepackage{amsmath}
\usepackage{amssymb}
\usepackage{mathtools}
\usepackage{amsthm}
\usepackage{xcolor}
\usepackage{makecell}
\usepackage[capitalize,noabbrev]{cleveref}
\crefname{lstlisting}{Listing}{Listings}
\usepackage{listings}
\usepackage[capitalize,noabbrev]{cleveref}
\crefname{lstlisting}{Listing}{Listings}
\usepackage{listings}
\usepackage[table]{xcolor}
\usepackage{tabularx}
\usepackage{ltablex}
\definecolor{rowblue}{HTML}{C9E5EE}
\usepackage{array}
\usepackage{caption}
\usepackage{enumitem}

\title{PrAg-PO: Prompt Augmented Policy Optimization for Robust and Diverse Mathematical Reasoning}

\author{%
  Wenquan Lu$^1$ \And Hai Huang$^2$ \And Enqi Liu$^3$ \And Randall Balestriero$^1$
}

\begin{document}

\maketitle
\vspace{-0.9cm}
\begin{center}
    \fontsize{9}{8}\selectfont
    \hspace{0.75em} $^1$Brown University \hspace{0.75em}  $^2$Palona AI\hspace{0.75em}$^3$Harvard University\\
    \texttt{\{wenquan\_lu, randall\_balestriero\}@brown.edu} \\ \texttt{hai@palona.ai} \hspace{0.75em}\texttt{enqi\textunderscore liu@hsph.harvard.edu}  \\
\end{center}

\begin{abstract}
Reinforcement learning algorithms such as group-relative policy optimization (GRPO) have shown strong potential for improving the mathematical reasoning capabilities of large language models. While a growing body of work seeks to improve training entropy, rollout diversity, and exploration, most existing methods still train models with a single fixed reasoning prompt or template, which can encourage prompt-specific overfitting and unstable training dynamics. In this work, we introduce Prompt Augmented Policy Optimization (PrAg-PO), a simple policy optimization method that mixes prompt templates with template-specific format rewards during training. By encouraging models to generate reasoning traces under diverse instructions and output formats, PrAg-PO increases rollout diversity and improves robustness. Compared with GRPO and DAPO, PrAg-PO achieves significantly higher reasoning accuracy while mitigating premature training collapse. Empirically, experiments on DeepSeek-R1-Distill-Qwen-1.5B, Qwen2.5-Math-1.5B, and Qwen3-1.7B  show that PrAg-PO consistently outperforms strong baselines and achieves competitive performance against recent methods on mathematics benchmarks, using only a fixed MATH Level 3-5 training set of 8.5K problems. The code and model checkpoints are available at \url{https://github.com/wenquanlu/PrAg-PO}.
\end{abstract}

\begin{figure}[H]
    \centering
    \includegraphics[width=\textwidth]{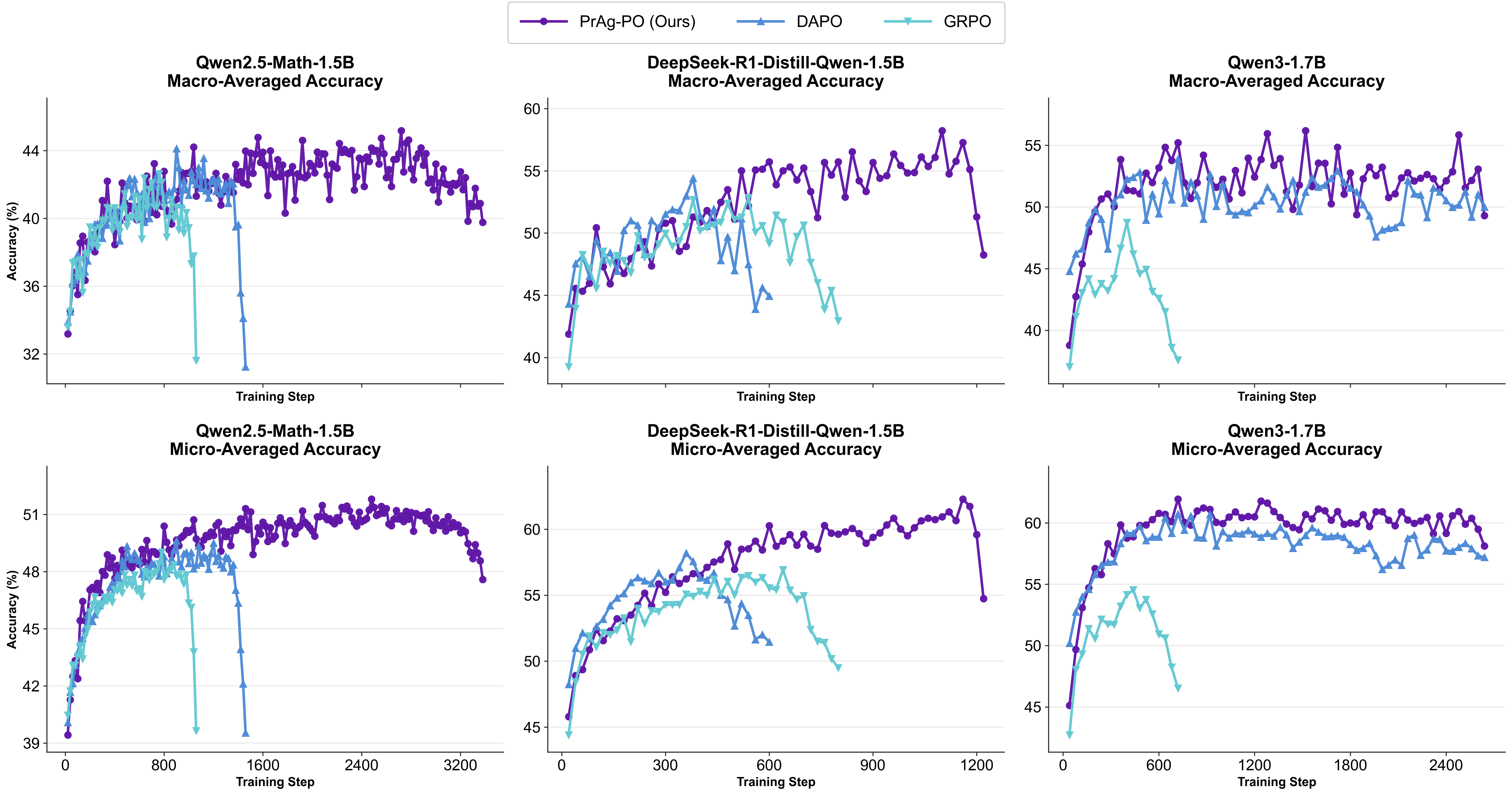}
    \caption{Evaluation accuracy over training steps averaged over mathematics benchmarks for models trained on MATH Level 3-5. PrAg-PO significantly outperforms GRPO and DAPO across Qwen2.5, DeepSeek-R1-Distilled and Qwen3 models, substantially delaying the overfitting and model collapse. }
    \label{fig:teaser}
\end{figure}

\section{Introduction}
\vspace{-5pt}
Since the introduction of GRPO~\cite{shao2024deepseekmath} and DeepSeek-R1~\cite{guo2025deepseek}, which are built upon classical reinforcement learning algorithms such as TRPO~\cite{schulman2015trust} and PPO~\cite{schulman2017proximal}, reinforcement learning (RL) has demonstrated strong capability in improving models’ reasoning performance on verifiable tasks, including mathematics~\cite{yu2025dapo, zeng2025simplerlzoo, zheng2025group, zhao2025geometric, gao2025soft}, coding~\cite{pourreza2025reasoningsql}, medical diagnosis~\cite{wang2025reinforcement}, and tabular reasoning~\cite{yang2025table}. RL has proven to be an effective post-training paradigm for large language models.

Numerous empirical and theoretical efforts have been made to improve the training objective of the GRPO algorithm. Notably, DAPO~\cite{yu2025dapo}, Dr. GRPO~\cite{liu2025understanding}, and GSPO~\cite{zheng2025group} unanimously remove the KL loss term from the original GRPO objective, arguing that it may limit reasoning performance by constraining the trained model to remain too close to the reference model. In addition, many works~\cite{cui2025entropy, cheng2025reasoning, shen2025entropy} have proposed a variety of entropy regularization techniques to address entropy collapse phenomenon: a sharp decrease in training entropy as reinforcement learning post-training progresses. Entropy serves as a proxy for policy stochasticity and output diversity. Viewing an LLM as a softmax policy over tokens, entropy collapse corresponds to probability mass concentrating on a narrow subset of tokens, severely limiting the diversity of sampled reasoning trajectories.

Despite the advancements to promote diversity, almost all existing work trains LLMs using a single reasoning format per training run, with a fixed set of format rewards. For example, DeepSeek-R1~\cite{guo2025deepseek} and Open-R1~\cite{openr1} adopt a tagged reasoning format (e.g., \textless think\textgreater...\textless/think\textgreater \textless answer\textgreater...\textless/answer\textgreater), while SimpleRL-Zoo~\cite{zeng2025simplerlzoo} and OAT-Zero~\cite{liu2025there} rely on Qwen’s original chain-of-thought prompting with free-form generation. We hypothesize that such homogeneous reasoning formats encourage overfitting to a single reasoning style, thereby reducing reasoning diversity. Moreover, given the expressive power of chain-of-thought reasoning, there exists a large and largely unexplored design space in which models can reason in fundamentally different ways. Also, given LLM's notorious sensitivity to prompts~\cite{yang2024large, NEURIPS2025_lu}, optimizing across different prompt formats can improve model's robustness to prompt variations, thereby potentially improve generalization and test-time performance.

In this work, we introduce PrAg-PO, a simple yet effective policy optimization technique for RL post-training of LLMs on mathematical reasoning. Specifically, we mix multiple reasoning templates and formats within a single training run, including tagged reasoning–answer separation, free-form generation, explicit chain-of-thought prompting, and reflection-based formats. These templates are paired with template-specific format rewards to ensure faithful adherence during training. We show that PrAg-PO successfully elicits diverse reasoning behaviors within a single model trained in a single run. Importantly, under a fixed training dataset and the default evaluation prompt template, PrAg-PO achieves competitive performance on mathematical reasoning benchmarks. As shown in \cref{fig:teaser}, our method outperforms both vanilla GRPO and DAPO in terms of per-benchmark mean accuracy and per-question mean accuracy on Qwen2.5-Math-1.5B, DeepSeek-R1-Distill-Qwen-1.5B and Qwen3-1.7B. It is important to note that such improvement is non-trivial, as mixed template training inherently causes distributional difference between train and test time, potentially impairing the performance. We also find that prompt augmentation helps stabilizing RL training, reducing overfitting and allowing continued policy improvement. In summary, our contributions are three-fold:

1. We propose Prompt Augmented Policy Optimization (PrAg-PO) for RL-based mathematical reasoning, a simple, effective, and computationally inexpensive policy optimization method that employs a diverse set of reasoning templates with associated format rewards to train models to reason. We show that this approach elicits a richer set of reasoning trajectories.

2. Empirically, prompt augmentation stabilizes training dynamics and substantially delays model collapse, enabling longer training horizons. We further interpret prompt augmentation as a gradient regularizer that smooths optimization and reduces overfitting to a single prompt-reward interface.

3. Across different model families, PrAg-PO significantly outperforms GRPO and DAPO, achieving competitive performance that exceeds leading results from prior literature on standard mathematical reasoning benchmarks, as shown in \cref{tab:main_result_qwen2.5}, \cref{tab:main_result_deepseek}, and \cref{tab:main_result_qwen3}.

\vspace{5pt}
\section{Related Works}
\textbf{Reinforcement Learning with Verifiable Rewards.} DeepSeek-R1 shows that rule-based verifiable rewards using GRPO algorithm can significantly improve the reasoning capability of pre-trained LLMs. This leads to a surge in a new research paradigm. DAPO~\cite{yu2025dapo} proposes decoupled-clip, dynamic sampling, dropping KL penalty and token-level gradient loss that promotes exploration diversity. Dr. GRPO~\cite{liu2025understanding} identifies a length bias in original GRPO's objective and proposes removing normalization term. GSPO~\cite{zheng2025group} proposed sequence-level importance ratios which notably stabilize RL training for MoE models. GMPO~\cite{zhao2025geometric} optimizes geometric-mean instead of the arithmetic-mean of token-level rewards. MO-GRPO~\cite{ichihara2025mo} and GDPO~\cite{liu2026gdpo} proposes to decouple the normalization of individual rewards in multi-reward settings.
Overall, most research centers on enhancing the stability and performance of RLVR algorithms.

\textbf{Increasing Data and Rollout Diversity for RL.} Early seminal works in RL~\cite{williams1991function} have added an entropy regularization term to the policy objective to encourage exploration~\cite{mnih2016asynchronous, schulman2017proximal, haarnoja2018soft}. However, it does not result in notable performance gains for LLMs~\cite{cui2025entropy} due to LLM’s extremely large response space~\cite{shen2025entropy}, which has led to a growing body of work on entropy regularization~\cite{cui2025entropy, cheng2025reasoning, shen2025entropy}. Notably, DRA-GRPO~\cite{chen2025dra} proposes a diversity-aware reward to increase rollout diversity by leveraging submodular mutual information. Beyond-Pass@1~\cite{liang2025beyond} shows that problem synthesis and self-play can maintain policy entropy during training. Recent or concurrent works~\cite{li2025questa,dai2026harderl, le2026transform} also propose question augmentation, which augments question content via partial answers or reformulations. However, these methods do not focus on augmenting prompt templates. Moreover, question-level augmentation is typically more expensive and requires carefully designed curricula. In contrast, our method is substantially lighter and easier to reproduce.

\textbf{LLM's Notorious Sensitivity to Prompts.} LLM's performance has been shown to be notoriously sensitive to the prompts. Chain-of-Thought (CoT) prompting has been shown to substantially improve LLM reasoning performance~\cite{wei2022chain, lu2025latent}. \cite{wei2022chain} demonstrate few-shot CoT demonstrations enable complex reasoning, while \cite{kojima2022large} show that LLMs can perform zero-shot reasoning by simply prepending “Let’s think step by step” to the prompt. DeepSeek-R1~\cite{guo2025deepseek} adopts zero-shot prompting and explicitly reports that few-shot prompting consistently degrades performance, indicating that RL-trained models are highly sensitive to evaluation prompts. In contrast, Dr. GRPO~\cite{liu2025understanding} shows that RL training itself is sensitive to the choice of prompt templates. Large language models as optimizer~\cite{yang2024large} show that one can improve LLM's performance on mathematical reasoning simply by optimizing test-time prompts.

\section{PrAg-PO: Prompt Augmented Policy Optimization}
\label{sec:method}
\subsection{Preliminary: Group-Relative Policy Optimization}
Group Relative Policy Optimization (GRPO) is a variant of proximal policy optimization (PPO) algorithm that replaces value-function baselines with group-relative normalization. For each question $q$ drawn from a training distribution, GRPO samples a group of rollouts $\{o_1, o_2, ..., o_G\}$ from the old policy $\pi_{\theta_{\mathrm{old}}}$, and maximizes the following surrogate objective:

\begin{align}
&\mathcal{L}_{\mathrm{GRPO_{token}}}(\theta)
=
\mathbb{E}_{q \sim \mathcal{D}, \{o_i\}_{i=1}^G \sim \pi_{\theta_\mathrm{old}}} 
\Bigg[
\frac{1}{\sum_{i=1}^{G}|o_i|}
\sum_{i=1}^{G}\sum_{t=1}^{|o_i|}  \\ &
\left(\min\!\Big(
r_{i,t}(\theta)\hat{A}_{i,t},\;
\mathrm{clip}\big(r_{i,t}(\theta), 1-\epsilon, 1+\epsilon\big)\,\hat{A}_{i,t} 
\Big) - \beta D_{KL}^{i,t}\right)
\Bigg],
\end{align}

Where $$r_{i, t}(\theta) = \frac{\pi_\theta(o_{i,t}|q, o_{i, <t})}{\pi_{\theta_\mathrm{old}}(o_{i,t}|q, o_{i, <t})}, \hat{A}_{i,t} = \frac{R_i - \mathrm{mean}(\{R_i\}_{i=1}^G)}{\mathrm{std}(\{R_i\}_{i=1}^G)}$$

Instead of using sample-level loss proposed in the original GRPO formulation, we here express token-level policy-gradient loss, which is widely adopted in most RL libraries (e.g., TRL~\cite{vonwerra2020trl}, verl~\cite{sheng2024hybridflow} and slime~\cite{slime_github}), leading to fairer credit assignment especially in long-CoT scenarios.

\begin{figure}[t]
    \centering
    \includegraphics[width=\textwidth]{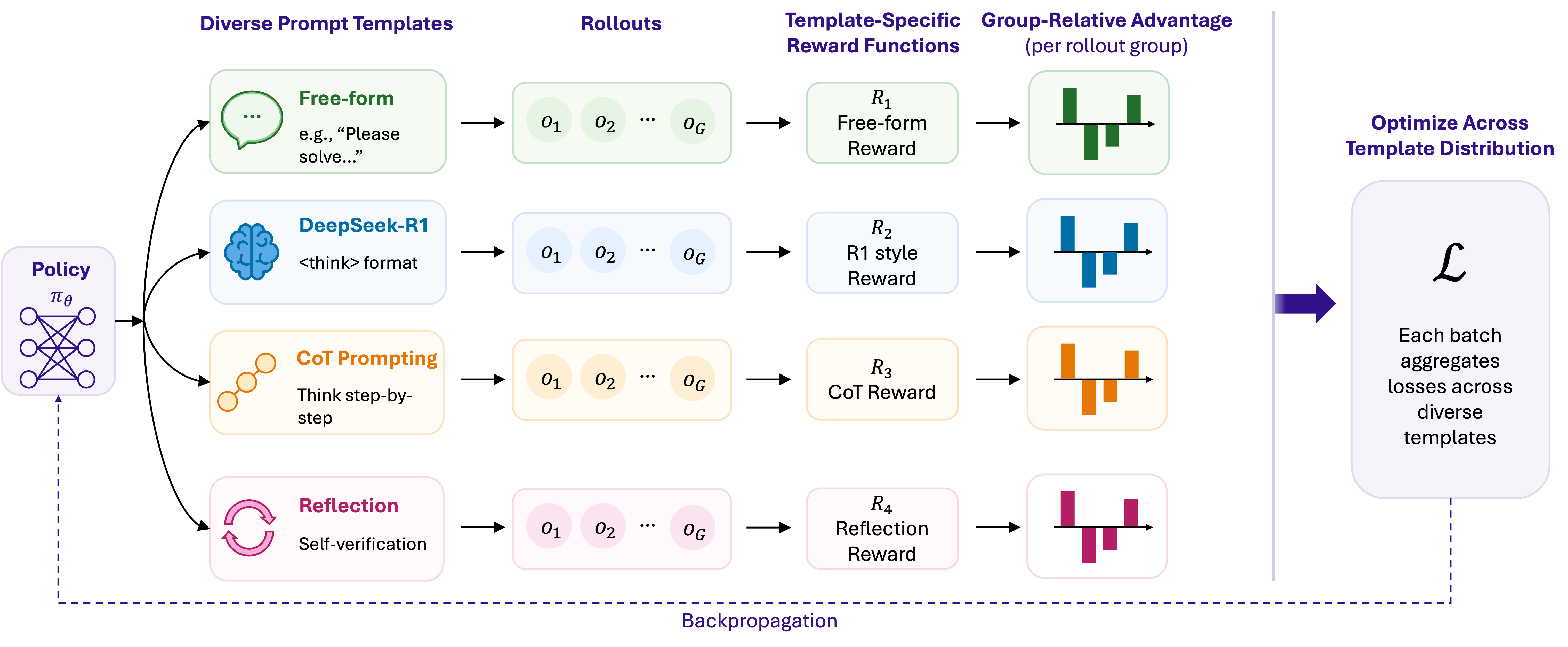}
    \caption{Outline of Prompt Augmented Policy Optimization. The policy is conditioned on diverse prompt templates to generate grouped rollouts. For each group, a template-specific reward function is applied and group-relative advantages are computed. Each batch update therefore aggregates losses across diverse prompt templates.}
    \label{fig:main_arch}
\end{figure}

\subsection{Prompt Augmentation}\label{sec:prompt_aug}
Here we introduce the formulation of prompt augmented policy optimization. Given a set of template functions $\{T_1, T_2, ..., T_K\}$ and corresponding reward functions $\{R_1, R_2, ..., R_K \}$ , each template function contains unique prompts and instructions that instruct the model to reason in a certain format. At training time, for each question $q$ drawn from a training distribution, we also uniformly sample a template function and apply it to the question $q$. We thus maximize the following objective:

\begin{align}
 &\mathcal{L}(\theta)
=
\mathbb{E}_{q \sim \mathcal{D}, \{o_i\}_{i=1}^G \sim \pi_{\theta_\mathrm{old}}, \color{red}{k\sim\mathrm{Uniform}(\{1,...K\})}} 
\Bigg[
\frac{1}{\sum_{i=1}^{G}|o_i|} 
\sum_{i=1}^{G}\\ &\sum_{t=1}^{|o_i|} 
\min\!\Big(
r_{i,t}^{{\color{red}{k}}}(\theta)\hat{A}_{i,t}^{{\color{red}{k}}},\;
\mathrm{clip}\big(r_{i,t}^{{\color{red}{k}}}(\theta), 1-\epsilon_\mathrm{low}, 1+\epsilon_\mathrm{high}\big)\,\hat{A}_{i,t}^{{\color{red}{k}}}
\Big)
\Bigg],
\end{align}

where \begin{align*}&r_{i, t}^{{\color{red}{k}}}(\theta) = \frac{\pi_\theta(o_{i,t}|{\color{red}{T_k(q)}}, o_{i, <t})}{\pi_{\theta_\mathrm{old}}(o_{i,t}|{\color{red}{T_k(q)}}, o_{i, <t})}, &\hat{A}_{i,t}^{{\color{red}{k}}} = \frac{{\color{red}{R_k}}(o_i) - \mathrm{mean}(\{{\color{red}{R_k}}(o_i)\}_{i=1}^G)}{\mathrm{std}(\{{\color{red}{R_k}}(o_i)\}_{i=1}^G)}\end{align*}

Note that we drop the KL penalty as a common practice~\cite{yu2025dapo, liu2025understanding}, and use decoupled clipping for entropy regularization. Using the above objective with prompt augmentation, we can elicit the language models to reason in diverse formats in a single training run. Moreover, each group of rollouts has the same template, which allows valid computation of advantage. As each batch contains multiple groups, each update still incorporates gradients from different templates, as shown in~\cref{fig:main_arch}. 

\begin{figure}[t]
    \centering
    \includegraphics[width=\textwidth]{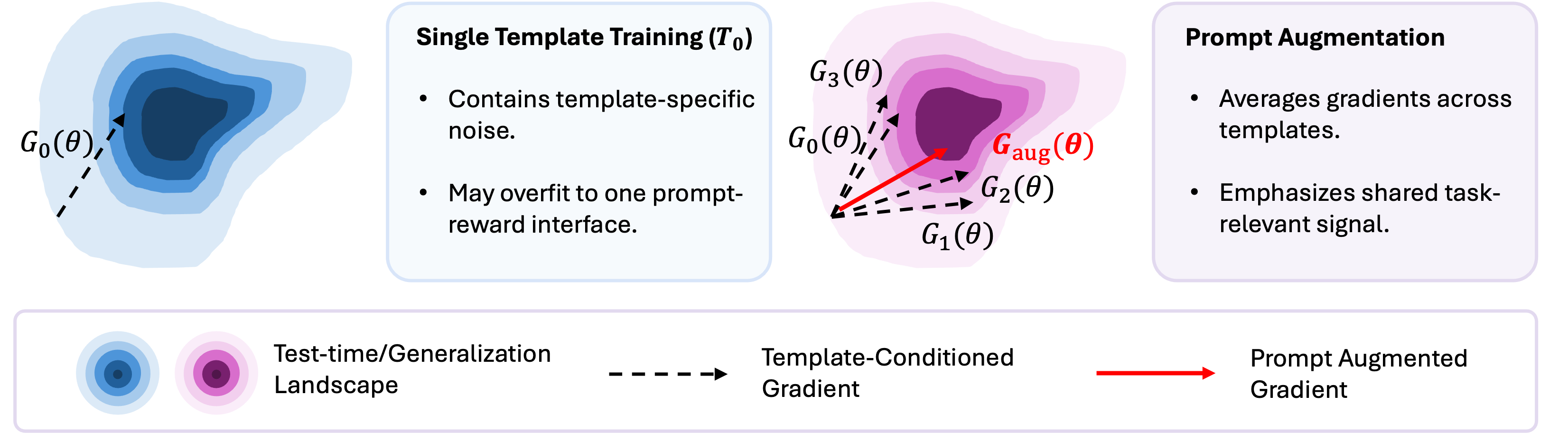}
    \caption{\textbf{Prompt augmentation as a gradient regularizer.} Left: Single-template training follows a template-conditioned gradient $G_0(\theta)$, which includes template-specific noise and can lead to overfitting to one prompt–reward interface. Right: Prompt augmentation samples multiple templates and averages their gradients $G_{\mathrm{aug}}(\theta) = \mathbb{E}_{k\sim P(T)}[G_k(\theta)]$, attenuating template-specific components while preserving shared task-relevant signals, resulting in  improved generalization.}
    \label{fig:analogy}
    \vspace{-10pt}
\end{figure}

\textbf{Optimization View: Prompt Augmentation as Gradient Regularizer.}
From an optimization perspective, prompt augmentation can be viewed as optimizing a prompt-perturbed training objective that regularizes policy optimization under the fixed training question distribution. Let \(T_k\) denote a prompt template sampled from \(P(T)\), where \(P=\mathrm{Uniform}(\{1,\ldots,K\})\). For each template, define the template-conditioned population gradient as
\[
G_k(\theta)
=
\mathbb{E}_{q\sim\mathcal{D}}
\left[
\nabla_\theta \mathcal{L}_{T_k}(q;\theta)
\right].
\]
PrAg-PO optimizes the augmented objective whose population gradient is
\[
G_{\mathrm{aug}}(\theta)
=
\mathbb{E}_{k\sim P(T)}[G_k(\theta)].
\]

Although evaluation is conducted under a single default prompt template, optimizing the augmented objective can still improve the default-template policy by acting as a regularizer during training. To see this, decompose the template-conditioned population gradient as
\[
G_k(\theta)
=
G_{\mathrm{shared}}(\theta)
+
\xi_{\mathrm{temp}}(k;\theta),
\]
where \(G_{\mathrm{shared}}(\theta)\) denotes gradient components that consistently improve mathematical reasoning across prompt-reward interfaces, while \(\xi_{\mathrm{temp}}(k;\theta)\) captures template-specific components induced by prompt wording, output structure, and format-reward design.

As illustrated in \cref{fig:analogy}, a single-template method repeatedly follows \(G_0(\theta)\), and may therefore over-optimize template-specific components associated with the default training prompt. Such components can reduce stability and lead to brittle reasoning behavior. In contrast, PrAg-PO averages updates over multiple prompt-reward interfaces. Under the idealized case where template-specific components are zero-mean across the prompt distribution,
$
\mathbb{E}_{k\sim P(T)}[\xi_{\mathrm{temp}}(k;\theta)] \approx 0,
$
the augmented gradient reduces to the shared reasoning gradient,
$G_{\mathrm{aug}}(\theta) \approx G_{\mathrm{shared}}(\theta).$ In practice, this cancellation need not be exact. As long as template-specific components are less consistently aligned across templates than the shared reasoning components, averaging over prompt-reward interfaces attenuates prompt-idiosyncratic directions relative to a single-template update. Thus, the prompt-augmented objective acts as a training-time regularizer, promoting more robust and generalizable gradients. This is consistent with our empirical results in \cref{fig:teaser}, where PrAg-PO delays training collapse, and with the entropy analysis in Appendix \ref{sec:entropy}, where it exhibits smoother token-level policy entropy.

\subsection{Prompt Template Curation and Selection}
\label{sec:curation}
We curate a diverse range of reasoning templates and formats to encourage the model to reason differently. Appendix~\ref{sec:all_prompts} illustrates examples of templates. Generally, the templates can be divided into four categories. The first group is the DeepSeek Style template that enforces the model to put reasoning processes within special tags like \textless think\textgreater \ tags and \textless answer\textgreater \ tags. Such templates are widely used in current literature and open-source projects~\cite{openr1, yang2025table}. The second group is the free-form reasoning generation, which is Qwen’s default style template. The model is instructed to output reasoning processes, and no format constraints are imposed except the final answer format. This style of template is also widely adopted in Qwen native training scenarios~\cite{zeng2025simplerlzoo, liu2025understanding}. The third group is reflection-based template. The model is instructed to check its work, identify any potential error, and put the verification processes within \textless check\textgreater\ tags. This form of reasoning is relatively underexplored in current literature, as most work investigated the natural emergence of "aha", "wait" or reflection moments rather than enforce an explicit format for reflection. The fourth group conditions the model on explicit chain-of-thought prompting. The models are conditioned on phrases like "Let's think step by step", and then continue the generation. For some templates with special tags, we additionally include a teacher-forced variant, where the assistant generation is initialized with the corresponding reasoning tags (e.g., \textless think\textgreater \ or \textless solution\textgreater). This provides stronger structural guidance during training and improves convergence speed.

Initially, we collect 13 prompt templates from open-source models and evaluation frameworks, including Qwen~\cite{yang2024qwen2}, DeepSeek-R1~\cite{guo2025deepseek}, Open-R1~\cite{openr1}, and lm-evaluation-harness~\cite{eval-harness}. We use these templates to train Qwen2.5-Math-1.5B and Qwen3-1.7B. In addition, we find that DeepSeek-R1-Distill-Qwen-1.5B benefits substantially from greater template diversity. Therefore, inspired by Toolformer~\cite{schick2023toolformer}, we leverage the in-context learning capability of a proprietary LLM~\cite{OpenAI_ChatGPT_2026} to expand the template pool to 35 prompt templates for training the distilled model, as detailed in Appendix~\ref{sec:template_deepseek}.

\begin{table}
\small
\centering
\setlength\tabcolsep{4.1pt}
\caption{Evaluation results on mathematics benchmarks for methods fine-tuned from Qwen2.5-Math-1.5B with reinforcement learning. See Appendix~\cref{tab:std_result1} for evaluation standard deviation.}
\renewcommand{\arraystretch}{1.2}
\begin{tabular}{l c c c c c | c c}
\hline
\textbf{Method} & \textbf{AIME24} & \textbf{AMC(23\&24)}  & \textbf{MATH500} & \textbf{Minerva} & \textbf{Olympiad} & \makecell[c]{\rule{0pt}{2.2ex}\textbf{AVG} \\ \scriptsize benchmark} & \makecell[c]{\rule{0pt}{2.2ex}\textbf{AVG} \\ \scriptsize question} \\\hline

Qwen2.5-Math-1.5B  & 16.7 & 43.4 & 61.8 & 15.1 & 28.4 & 33.1 & 37.4 \\
GRPO  & \textbf{23.3} & 49.9 & 75.4 & 26.3 & 38.3 & 42.6 & 48.4 \\
Dr. GRPO  & 20.0 & 53.0 & 74.2 & 25.7 & 37.6 & 42.1 & 47.7 \\
SEED GRPO & \textbf{23.3} & 50.6 & 75.4 & 26.8 & \textbf{41.3} & 43.5 & 49.8 \\
GMPO & 20.0 & 53.0 & 77.6 & 30.1 & 38.7  & 43.9 & 50.1 \\
DAPO  & \textbf{23.3} & \textbf{54.9} & 76.9 & 26.0 & 39.4 & 44.1 & 49.6 \\
\rowcolor{rowblue}
PrAg-PO (Step 2720) & \textbf{23.3} & 53.5  & 78.4 & \textbf{31.3} & 39.4  & \textbf{45.2} & 50.9 \\
\rowcolor{rowblue}PrAg-PO (Step 2480)  & 20.0 & 49.4  & \textbf{80.4} & 29.7 & 41.2 & 44.1 & \textbf{51.8}
\\\hline
\end{tabular}
\vspace{-10pt}
\label{tab:main_result_qwen2.5}
\end{table}

\begin{table}
\small
\centering
\setlength\tabcolsep{2.75pt}
\caption{Evaluation results on mathematics benchmarks for methods fine-tuned from DeepSeek-R1-Distill-Qwen-1.5B with reinforcement learning. See Appendix~\cref{tab:std_result2} for standard deviation.}
\renewcommand{\arraystretch}{1.2}
\begin{tabular}{l c c c c c | c c}
\hline
\textbf{Method} & \textbf{AIME24} & \textbf{AMC23}  & \textbf{MATH500} & \textbf{Minerva} & \textbf{Olympiad} & \makecell[c]{\rule{0pt}{2.2ex}\textbf{AVG} \\ \scriptsize benchmark} & \makecell[c]{\rule{0pt}{2.2ex}\textbf{AVG} \\ \scriptsize question} \\\hline

DeepSeek-R1-Distill-Qwen-1.5B  & 28.8 & 62.9 & 82.8  & 26.5 & 43.3 & 48.9 & 53.5 \\
Still-3-1.5B-Preview & 32.5 & 66.7 & 84.4  & 29.0 & 45.4 & 51.6 & 55.6 \\
DeepScaleR-1.5B-Preview  & 43.1 & 73.6 & \textbf{87.8}  & 30.2 & 50.0 & 57.0 & 59.4 \\
Open-RS1 & 30.0 & 70.0 & 83.8 & 29.0 & 52.4 & 53.0 & 58.6 \\
Open-RS2 & 30.0 & 80.0 & 85.4 & 30.5 & 52.4 & 55.7 & 59.6 \\
Open-RS3  & \textbf{46.7} & 72.5 & 84.4 & 26.8 & 51.3 & 56.3 & 58.3 \\ 
Dr. GRPO & 33.3 & 80.0 & 83.4 & 30.5 & 52.1 & 56.0 & 58.9 \\
DRA-GRPO & 36.7 & 75.0 & 86.2 & 32.4 & 53.0 & 56.7 & 60.5 \\
DRA-Dr.GRPO  & 36.7 & \textbf{85.0} & 85.2 & 30.5 & 53.8 & \textbf{58.2} & 60.5 \\
GRPO & 30.0 & 74.0 & 85.5 & 28.2 & 46.5 & 52.8 & 56.5  \\
DAPO & 33.3 & 77.5 & 84.2 & 26.9 & 50.0 & 54.4 & 57.6 \\
\rowcolor{rowblue}
PrAg-PO (Step 1100) & 42.0 & 78.5 & 87.5 & 29.3 & 53.8 & \textbf{58.2} & 60.9 \\
\rowcolor{rowblue}
PrAg-PO (Step 1160) & 33.3 & 77.0 & 87.2 & \textbf{32.6} & \textbf{56.2} & 57.3 & \textbf{62.3}
\\\hline
\end{tabular}
\label{tab:main_result_deepseek}
\vspace{-10pt}
\end{table}

\begin{table}[H]
\setlength\tabcolsep{5.4pt}
\small
\centering
\caption{Evaluation results on mathematics benchmarks for methods fine-tuned from Qwen3-1.7B with reinforcement learning. See Appendix~\cref{tab:std_result3} for evaluation standard deviation.}
\renewcommand{\arraystretch}{1.2}
\begin{tabular}{l c c c c c | c c}
\hline
\textbf{Method} & \textbf{AIME24} & \textbf{AMC23}  & \textbf{MATH500} & \textbf{Minerva} & \textbf{Olympiad} & \makecell[c]{\rule{0pt}{2.2ex}\textbf{AVG} \\ \scriptsize benchmark} & \makecell[c]{\rule{0pt}{2.2ex}\textbf{AVG} \\ \scriptsize question} \\\hline
Qwen3-1.7B & 3.3 & 27.5 & 57.9 & 19.1 & 18.2 & 25.2 & 32.9 \\
GAC & 20.0 & 54.1 & 73.6 & 25.1 & 28.1 & 40.2 & 43.1 \\
RePO & 12.3 & 47.7 & 59.6 & 20.2 & \textbf{63.1} & 40.6 & 52.8 \\
GRPO & 16.7 & 70.0 & 78.7 & 33.2 & 45.1 & 48.7 & 54.1 \\
DAPO & 26.7 & 70.0 & 84.9 & 32.8 & 55.1 & 53.9 & 60.7 \\
\rowcolor{rowblue}
PrAg-PO (Step 1520)  & \textbf{33.3} & 75.0 & 85.1 & \textbf{33.6} & 53.9 & \textbf{56.2} & 60.7 \\
\rowcolor{rowblue}
PrAg-PO (Step 720)  & 23.3 & \textbf{76.5} & \textbf{87.0} & 33.5 & 55.7 & 55.2 & \textbf{62.0}
\\\hline
\end{tabular}
\label{tab:main_result_qwen3}
\vspace{-10pt}
\end{table}

\subsection{Template-Specific Reward Functions}
The reward for PrAg-PO is composed of an accuracy reward and a format reward, both ranging from 0 to 1. We use binary rewards for accuracy, and follow TreePO~\cite{li2025treepo}, using Math-Verify~\cite{Kydlicek_MathVerify_2024} and SymPy~\cite{10.7717/peerj-cs.103} libraries to verify a response by extracting the final answer enclosed within \texttt{\textbackslash boxed\{\}}.

\textbf{Template-Specific reward functions are critical to elicit diverse reasoning formats from the model.} We will show in ablation studies that without these specific reward functions, the model will simply ignore the instruction and still collapse prematurely. Nevertheless, we found that simple string matching and tag-count-based rewards are sufficient to induce the desired reasoning format, without requiring complex regular expressions (Regex) matching. As shown in Appendix~\ref{sec:all_prompts}, for DeepSeek-style format, format rewards are computed based on the exact match of special tags. Partial rewards are assigned when the model generates a subset of the required tags; missing or duplicated tags receive no reward. Moreover, the ordering of the tags are not examined, as empirically the model can naturally develop correct ordering. The observation that language models are quick format learners are also observed in other literature~\cite{xie2025interleaved}, however, we are the first to formally explore it in a hybrid format setting. For templates that do not impose format constraints, we assign the format reward a constant value of 1 to ensure a consistent scale across templates. Since the GRPO objective normalizes rewards within each group, this constant has no effect on the advantage estimates.

\vspace{-5pt}
\section{Experiments}
\vspace{-5pt}
\label{sec:main_exp}
\subsection{Experimental Settings}
\vspace{-5pt}
\label{sec:settings}
\textbf{Datasets.} We use the MATH Level 3-5 dataset~\cite{zeng2025simplerlzoo} as the training set, which is a subset of 8,523 questions derived from the MATH dataset~\cite{hendrycks2measuring}. This filtered subset contains problems of medium to high difficulty (i.e., levels 3 to 5 on a five-level scale). The MATH dataset consists of problems from mathematics competitions, including AMC 10, AMC 12, and AIME. Our test set comprises a standard suite of benchmarks, including AIME24 (30 questions)~\cite{aime24}, AMC (83 questions)~\cite{aimo2024validationamc}, AMC23 (40 questions), MATH500 (500 questions), Minerva Math (272 questions)~\cite{lewkowycz2022solving}, and OlympiadBench (675 questions)~\cite{he2024olympiadbench}, covering a broad range of graduate-level and Olympiad-style problems.

\textbf{Models and Competitors.}
We use Qwen2.5-Math-1.5B~\cite{yang2024qwen2}, DeepSeek-R1-Distill-Qwen-1.5B~\cite{guo2025deepseek}, and Qwen3-1.7B~\cite{qwen3technicalreport} as base models, covering different model families and training stages. Qwen2.5-Math-1.5B is continually pretrained from Qwen2.5-1.5B on mathematical corpora. DeepSeek-R1-Distill-Qwen-1.5B is distilled from the large DeepSeek-R1 model. Qwen3-1.7B is a latest-generation instruction-tuned model.

We reproduce GRPO~\cite{shao2024deepseekmath} and DAPO~\cite{yu2025dapo} baselines using our code base. For GRPO, we adopt token-level loss. KL coefficient $\beta$ is 0.001, consistent with DeepSeek-R1. For DAPO, we implement decoupled clip and KL penalty removal, following the core training objective. In addition to GRPO and DAPO, we compare our method against a comprehensive set of state-of-the-art baselines. For Qwen2.5-Math-1.5B, we include Dr. GRPO~\cite{liu2025understanding}, SEED-GRPO~\cite{chen2025seed}, and GMPO~\cite{zhao2025geometric}. For DeepSeek-R1-Distill-Qwen-1.5B, we include Still-3-1.5B-Preview~\cite{Slow_Thinking_with_LLMs_3_Preview}, DeepScaleR-1.5B-Preview~\cite{deepscaler2025}, Open-RS~\cite{dang2026reinforcement}, Dr. GRPO~\cite{liu2025understanding}, and DRA-GRPO~\cite{chen2025dra}. For Qwen3-1.7B, we include GAC~\cite{xu2026gac} and RePO~\cite{li2025repo}.

\textbf{Training and Evaluation Settings.} We train our models using the verl framework~\cite{sheng2024hybridflow}. We use a prompt batch size of 128 with a mini-batch size of 32, resulting in four weight updates per batch. The group size is set to 8, so each mini-batch contains 256 rollouts. Following DAPO, we set the clipping parameters to $\epsilon_{\mathrm{high}} = 0.28$ and $\epsilon_{\mathrm{low}} = 0.20$. For Qwen2.5-Math-1.5B and Qwen3-1.7B, we cap the maximum prompt length at 1024 tokens and the maximum output length at 3072 tokens during training, and use a maximum model length of 4096 tokens during evaluation. For DeepSeek-R1-Distill-Qwen-1.5B, following DRA-GRPO~\cite{chen2025dra}, we cap the maximum prompt length at 1024 tokens and the maximum output length at 3584 tokens during training, and use a maximum model length of 32768 tokens during evaluation. We use AdamW optimizer and a constant learning rate of $1\times10^{-6}$. All training runs are conducted on 8 L40S GPUs, with vLLM serving as the inference engine. Each training run takes approximately 1.5 - 4 days to complete depending on the total optimization steps.

\label{sec:eval_setting} To ensure fair comparison and avoid introducing test-time scaling effects, \textbf{we use model tokenizer's single default template during evaluation,} adopting exactly the same evaluation framework and code base as Dr. GRPO, SEED GRPO and GMPO. We evaluate models' performance under greedy decoding (i.e., temperature = 0), and report pass@1 metric. Furthermore, due to inherent stochasticity of vLLM inference engine~\cite{kwon2023efficient}, the model may produce different responses even under the same seed. To ensure reproducibility and reliability of our results, we let all models perform five inference rounds under the same seed, and report the average accuracy. Importantly, due to the extreme imbalanced nature of the test datasets, we report both per-benchmark average accuracy and per-question average accuracy. Reporting both metrics avoids bias toward either large benchmarks  or small benchmarks in a highly imbalanced test suite. Following standard practice in Dr. GRPO and DAPO, the model is evaluated at periodic checkpoints, and the best-performing results are reported in the table.

\subsection{PrAg-PO Improves Reasoning Accuracy and Stabilizes Group-Relative Policy Training}

PrAg-PO demonstrates strong performance in both internal comparisons against our GRPO and DAPO baselines and external comparisons against previously reported methods. As shown in \cref{tab:main_result_qwen2.5}, \cref{tab:main_result_deepseek}, and \cref{tab:main_result_qwen3}, PrAg-PO consistently outperforms the GRPO and DAPO baselines in both average benchmark accuracy and average question accuracy, indicating that its performance gains are robust across evaluation criteria. Compared with other competitive and state-of-the-art methods, PrAg-PO also achieves leading performance across model families. For Qwen2.5-Math-1.5B, shown in \cref{tab:main_result_qwen2.5}, PrAg-PO achieves the best performance under both evaluation metrics. Notably at step 2480, to the best of our knowledge, this is the first instance where a Qwen2.5-Math-1.5B RL-finetuned model surpasses \textbf{80\%} accuracy on the MATH500 benchmark. For DeepSeek-R1-Distill-Qwen-1.5B, shown in \cref{tab:main_result_deepseek}, PrAg-PO again achieves the highest average accuracy in both metrics, having a 2-point lead in per-question accuracy, while also obtaining the best score on OlympiadBench and Minerva. For Qwen3-1.7B, shown in \cref{tab:main_result_qwen3}, PrAg-PO substantially outperforms all competing methods, exceeding the strongest previously reported result (i.e., RePO) by more than 10 points. These results demonstrate that prompt augmentation is an effective and broadly applicable method for improving the reasoning accuracy of language models under reinforcement learning fine-tuning.

Training collapse during prolonged GRPO training has been observed in numerous prior studies~\cite{dang2026reinforcement, qi2025precisionrl, liu-li-2025-rl-collapse, liu2025prorl}. Although its exact causes remain an active area of research, recent work has attributed such instability to factors such as entropy collapse and training--inference precision mismatch. Empirically, we find that PrAg-PO is effective in stabilizing training dynamics and delaying collapse. As shown in \cref{fig:teaser}, Dr. GRPO and DAPO collapse after roughly 400--1500 steps. In contrast, PrAg-PO continues to improve over a substantially longer training horizon, reaching more than 2000 steps. This effect is also reflected in token-level policy entropy, where PrAg-PO exhibits smoother and more stable entropy trajectories compared to DAPO (See Appendix~\ref{sec:entropy}). This improved stability may be driven by the regularizing effect of prompt augmentation. By exposing the policy to diverse prompt and reward interfaces, PrAg-PO reduces overfitting to any single reasoning template. Moreover, aggregating losses across different templates smooths the overall gradient signal, as illustrated in \cref{sec:prompt_aug}, attenuating template-specific noise and reducing abrupt spikes in training dynamics. 

\begin{figure}[!t]
\centering
\includegraphics[width=\textwidth]{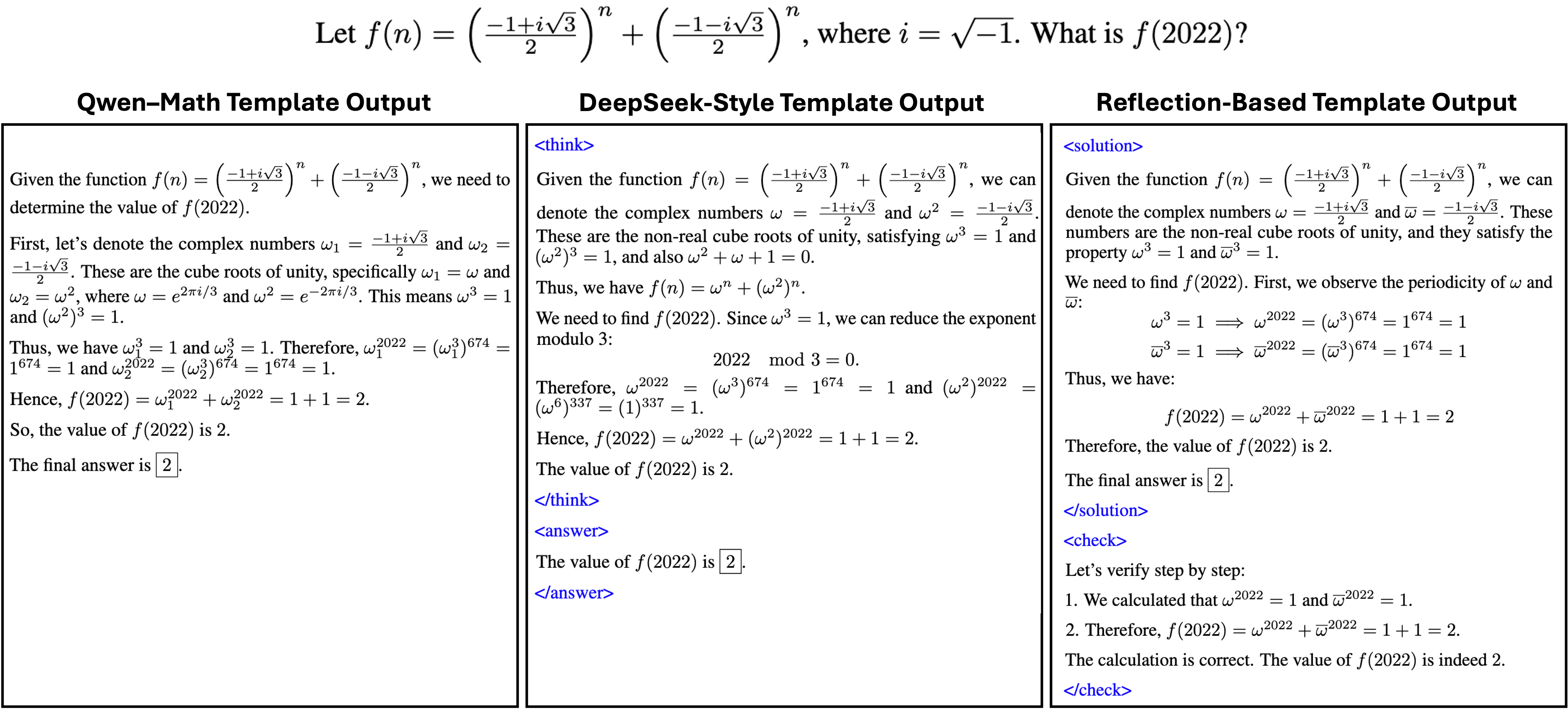}
\caption{Reasoning trajectories produced by the model trained with PrAg-PO when answering the same question under different prompt templates. The model generates diverse reasoning formats.}
\label{fig:example_output}
\vspace{-10pt}
\end{figure}

\begin{figure*}[t]
\centering

\begin{minipage}[t]{0.49\textwidth}
    \centering
    \includegraphics[width=\linewidth]{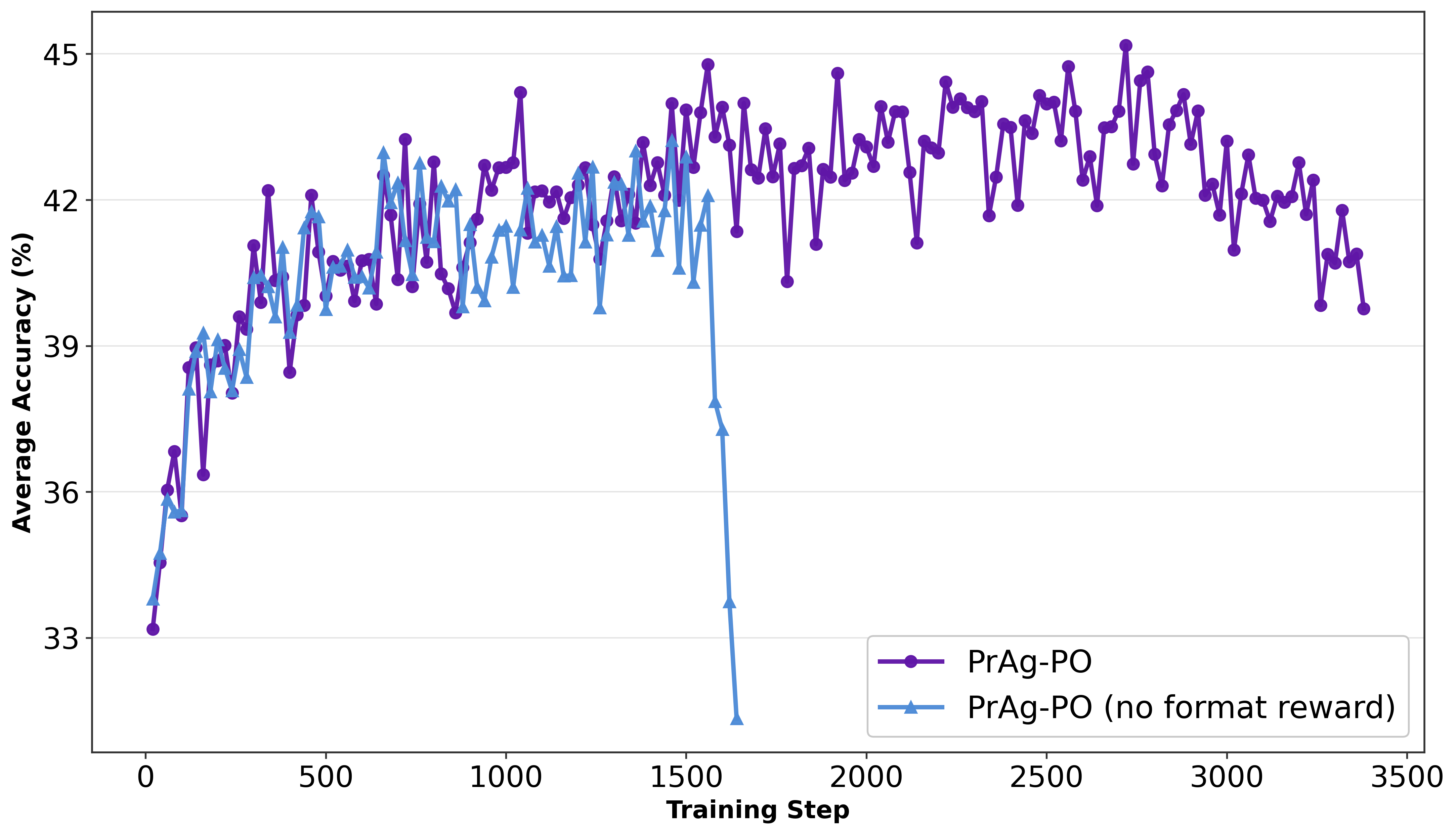}
    \captionof{figure}{Accuracy of PrAg-PO with and without format rewards across training steps on Qwen2.5-Math-1.5B. Without format rewards (blue curve), the model underperforms and collapses earlier.}
    \label{fig:no_format_acc}
\end{minipage}
\hfill
\begin{minipage}[t]{0.49\textwidth}
    \centering
    \includegraphics[width=\linewidth]{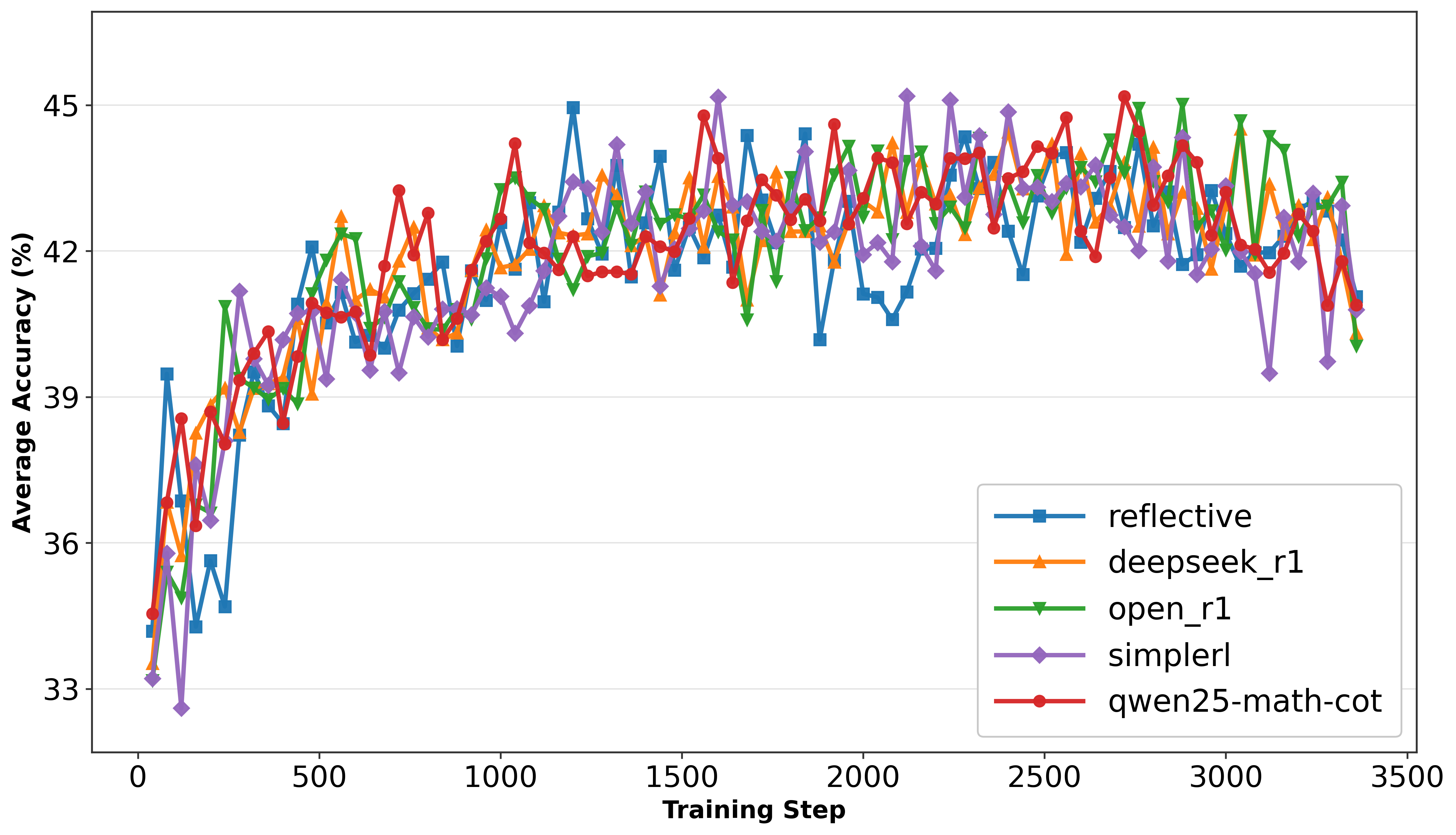}
    \captionof{figure}{Accuracy of PrAg-PO evaluated with different prompt templates on Qwen2.5-Math-1.5B. Performance is consistent and robust across prompt templates, with no clear outlier.}
    \label{fig:template_acc}
\end{minipage}
\vspace{-10pt}

\end{figure*}

\subsection{PrAg-PO Elicits Diverse Reasoning Trajectories}

In this section, we demonstrate that PrAg-PO enables the model to develop diverse reasoning formats and trajectories. As shown in \cref{fig:example_output}, we evaluate a single trained checkpoint on the same mathematical question, ``Let $f(n) = \left(\frac{-1 + i\sqrt{3}}{2}\right)^n + $...", while varying only the prompt template. Despite identical inputs and model parameters, the model adapts its output to follow the instructions imposed by each template, producing distinct reasoning structures. In the left column, the Qwen-Math free-form generation template is applied, resulting in a generic, unconstrained reasoning trace. In the middle column, a DeepSeek-style template instructs the model to place its reasoning within \textless think\textgreater tags and its final answer within \textless answer\textgreater tags. The model successfully adheres to these structural constraints, with the required tags highlighted in blue. In the right column, a reflection-based template is used, prompting the model to include an additional verification step enclosed in \textless check\textgreater tags, which it also follows correctly. Although the three solutions are mathematically similar, they exhibit subtle yet meaningful differences in reasoning emphasis and presentation. For instance, the left-column output explicitly converts the complex numbers into their exponential form to invoke the cube roots of unity, whereas the middle-column output emphasizes reducing the exponent $2022 \ \mathrm{ modulo } \ 3$ to simplify the expression. Interestingly, the right-column output first identifies $\frac{-1 - i\sqrt{3}}{2}$ as the complex conjugate of $\frac{-1 + i\sqrt{3}}{2}$ before appealing to the same cube-root-of-unity property, and presents the solution in a more step-wise manner that facilitates verification. These observations illustrate that prompt augmentation elicits multiple valid reasoning trajectories from a single model, diversifying its outputs and supporting more effective GRPO training.

\vspace{-5pt}
\subsection{Ablation Studies}
\vspace{-5pt}
\textbf{Template-Specific Format Rewards Are Critical for Stable Training.} In this section, we examine whether template-specific format rewards are necessary for improving reasoning performance and eliciting diverse reasoning trajectories. In principle, these benefits could be attributed solely to prompt-level data augmentation, without modifying the training objective. To isolate the effect of format rewards, we remove them entirely and train the model using only a binary accuracy reward, while still applying the augmented prompt templates. The resulting accuracy curves are shown as the light blue line in \cref{fig:no_format_acc}. Without format rewards, training collapses at approximately 1,500 steps, and the model consistently underperforms the counterpart trained with format rewards. Through manual inspection of the model’s rollouts, we observe that under this setting the model largely ignores the prompt instructions and defaults to free-form reasoning outputs. This behavior is expected, as in the absence of explicit reward signals, the model has no incentive to follow diverse reasoning formats.

\textbf{There is No ``Secret'' Gold Prompt.}
A natural concern is that the effectiveness of PrAg-PO may be driven by the accidental inclusion of a ``gold'' prompt that substantially boosts training performance. In this section, we show that this is not the case: after prompt-augmented policy optimization, the model performs comparably across different prompt templates, indicating that no single template provides a disproportionate advantage. We evaluate Qwen2.5-Math-1.5B at periodic checkpoints under different prompt templates. As shown in \cref{fig:template_acc}, accuracies are largely consistent across templates, fluctuating around a common mean without clear outliers. This suggests that the model becomes robust to prompt variation, and that the performance gains arise from the collective effect of prompt augmentation rather than from a single hidden ``gold'' prompt.

\vspace{-5pt}
\section{Conclusion, Limitations and Future Works}
\label{sec:conclusion}
\vspace{-5pt}
We introduce Prompt Augmented Policy Optimization (PrAg-PO) for training LLMs on mathematical reasoning tasks. We show that prompt augmentation, when combined with template-specific format rewards, elicits diverse reasoning trajectories, stabilizes training, and improves reasoning accuracy. One limitation is that PrAg-PO does not fully eliminate long-horizon optimization instability, as the model may still collapse after sufficiently prolonged training. Nevertheless, under a fixed dataset budget, PrAg-PO enables more stable training dynamics and achieves stronger performance. Another limitation is that the template pool still involves human design choices. More systematic template selection could improve scalability, although our analysis suggests that PrAg-PO does not rely on a single ``gold'' prompt. As a promising direction for future work, the availability of multiple reasoning formats opens the door to inference-time scaling strategies that may further enhance performance.

\bibliographystyle{plainnat}
\bibliography{references}

@article{shao2024deepseekmath,
  title={Deepseekmath: Pushing the limits of mathematical reasoning in open language models},
  author={Shao, Zhihong and Wang, Peiyi and Zhu, Qihao and Xu, Runxin and Song, Junxiao and Bi, Xiao and Zhang, Haowei and Zhang, Mingchuan and Li, YK and Wu, Yang and others},
  journal={arXiv preprint arXiv:2402.03300},
  year={2024}
}

@article{guo2025deepseek,
  title={Deepseek-r1: Incentivizing reasoning capability in llms via reinforcement learning},
  author={Guo, Daya and Yang, Dejian and Zhang, Haowei and Song, Junxiao and Zhang, Ruoyu and Xu, Runxin and Zhu, Qihao and Ma, Shirong and Wang, Peiyi and Bi, Xiao and others},
  journal={arXiv preprint arXiv:2501.12948},
  year={2025}
}

@inproceedings{schulman2015trust,
  title={Trust region policy optimization},
  author={Schulman, John and Levine, Sergey and Abbeel, Pieter and Jordan, Michael and Moritz, Philipp},
  booktitle={International conference on machine learning},
  pages={1889--1897},
  year={2015},
  organization={PMLR}
}

@article{schulman2017proximal,
  title={Proximal policy optimization algorithms},
  author={Schulman, John and Wolski, Filip and Dhariwal, Prafulla and Radford, Alec and Klimov, Oleg},
  journal={arXiv preprint arXiv:1707.06347},
  year={2017}
}

@inproceedings{yang2025table,
  title={Table-r1: Inference-time scaling for table reasoning tasks},
  author={Yang, Zheyuan and Chen, Lyuhao and Cohan, Arman and Zhao, Yilun},
  booktitle={Proceedings of the 2025 Conference on Empirical Methods in Natural Language Processing},
  pages={20616--20635},
  year={2025}
}

@article{cui2025entropy,
  title={The entropy mechanism of reinforcement learning for reasoning language models},
  author={Cui, Ganqu and Zhang, Yuchen and Chen, Jiacheng and Yuan, Lifan and Wang, Zhi and Zuo, Yuxin and Li, Haozhan and Fan, Yuchen and Chen, Huayu and Chen, Weize and others},
  journal={arXiv preprint arXiv:2505.22617},
  year={2025}
}

@article{cheng2025reasoning,
  title={Reasoning with exploration: An entropy perspective},
  author={Cheng, Daixuan and Huang, Shaohan and Zhu, Xuekai and Dai, Bo and Zhao, Wayne Xin and Zhang, Zhenliang and Wei, Furu},
  journal={arXiv preprint arXiv:2506.14758},
  year={2025}
}

@article{chen2025seed,
  title={Seed-grpo: Semantic entropy enhanced grpo for uncertainty-aware policy optimization},
  author={Chen, Minghan and Chen, Guikun and Wang, Wenguan and Yang, Yi},
  journal={arXiv preprint arXiv:2505.12346},
  year={2025}
}

@inproceedings{
zeng2025simplerlzoo,
title={Simple{RL}-Zoo: Investigating and Taming Zero Reinforcement Learning for Open Base Models in the Wild},
author={Weihao Zeng and Yuzhen Huang and Qian Liu and Wei Liu and Keqing He and Zejun MA and Junxian He},
booktitle={Second Conference on Language Modeling},
year={2025},
url={https://openreview.net/forum?id=vSMCBUgrQj}
}

@misc{openr1,
    title = {Open R1: A fully open reproduction of DeepSeek-R1},
    url = {https://github.com/huggingface/open-r1},
    author = {{Hugging Face}},
    month = {January},
    year = {2025}
}

@inproceedings{
yu2025dapo,
title={{DAPO}: An Open-Source {LLM} Reinforcement Learning System at Scale},
author={Qiying Yu and Zheng Zhang and Ruofei Zhu and Yufeng Yuan and Xiaochen Zuo and YuYue and Weinan Dai and Tiantian Fan and Gaohong Liu and Juncai Liu and LingJun Liu and Xin Liu and Haibin Lin and Zhiqi Lin and Bole Ma and Guangming Sheng and Yuxuan Tong and Chi Zhang and Mofan Zhang and Ru Zhang and Wang Zhang and Hang Zhu and Jinhua Zhu and Jiaze Chen and Jiangjie Chen and Chengyi Wang and Hongli Yu and Yuxuan Song and Xiangpeng Wei and Hao Zhou and Jingjing Liu and Wei-Ying Ma and Ya-Qin Zhang and Lin Yan and Yonghui Wu and Mingxuan Wang},
booktitle={The Thirty-ninth Annual Conference on Neural Information Processing Systems},
year={2025},
url={https://openreview.net/forum?id=2a36EMSSTp}
}

@inproceedings{
liu2025understanding,
title={Understanding R1-Zero-Like Training: A Critical Perspective},
author={Zichen Liu and Changyu Chen and Wenjun Li and Penghui Qi and Tianyu Pang and Chao Du and Wee Sun Lee and Min Lin},
booktitle={Second Conference on Language Modeling},
year={2025},
url={https://openreview.net/forum?id=5PAF7PAY2Y}
}

@article{zheng2025group,
  title={Group sequence policy optimization},
  author={Zheng, Chujie and Liu, Shixuan and Li, Mingze and Chen, Xiong-Hui and Yu, Bowen and Gao, Chang and Dang, Kai and Liu, Yuqiong and Men, Rui and Yang, An and others},
  journal={arXiv preprint arXiv:2507.18071},
  year={2025}
}

@article{shen2025entropy,
  title={On entropy control in llm-rl algorithms},
  author={Shen, Han},
  journal={arXiv preprint arXiv:2509.03493},
  year={2025}
}

@misc{liu2025there,
  title={There May Not be Aha Moment in R1-Zero-like Training — A Pilot Study},
  author={Liu, Zichen and Chen, Changyu and Li, Wenjun and Pang, Tianyu and Du, Chao and Lin, Min},
  year={2025},
  howpublished={\url{https://oatllm.notion.site/oat-zero}},
  note={Notion Blog},
}

@article{zhao2025geometric,
  title={Geometric-mean policy optimization},
  author={Zhao, Yuzhong and Liu, Yue and Liu, Junpeng and Chen, Jingye and Wu, Xun and Hao, Yaru and Lv, Tengchao and Huang, Shaohan and Cui, Lei and Ye, Qixiang and others},
  journal={arXiv preprint arXiv:2507.20673},
  year={2025}
}

@article{gao2025soft,
  title={Soft adaptive policy optimization},
  author={Gao, Chang and Zheng, Chujie and Chen, Xiong-Hui and Dang, Kai and Liu, Shixuan and Yu, Bowen and Yang, An and Bai, Shuai and Zhou, Jingren and Lin, Junyang},
  journal={arXiv preprint arXiv:2511.20347},
  year={2025}
}

@inproceedings{
pourreza2025reasoningsql,
title={Reasoning-{SQL}: Reinforcement Learning with {SQL} Tailored Partial Rewards for Reasoning-Enhanced Text-to-{SQL}},
author={Mohammadreza Pourreza and Shayan Talaei and Ruoxi Sun and Xingchen Wan and Hailong Li and Azalia Mirhoseini and Amin Saberi and Sercan O Arik},
booktitle={Second Conference on Language Modeling},
year={2025},
url={https://openreview.net/forum?id=HbwkIDWQgN}
}

@inproceedings{
wang2025reinforcement,
title={Reinforcement Learning for Out-of-Distribution Reasoning in {LLM}s: An Empirical Study on Diagnosis-Related Group Coding},
author={Hanyin Wang and Zhenbang Wu and Gururaj J. Kolar and Hariprasad Reddy Korsapati and Brian Bartlett and Bryan Hull and Jimeng Sun},
booktitle={The Thirty-ninth Annual Conference on Neural Information Processing Systems},
year={2025},
url={https://openreview.net/forum?id=0jvnfH0WYV}
}

@article{ichihara2025mo,
  title={Mo-grpo: Mitigating reward hacking of group relative policy optimization on multi-objective problems},
  author={Ichihara, Yuki and Jinnai, Yuu and Morimura, Tetsuro and Sakamoto, Mitsuki and Mitsuhashi, Ryota and Uchibe, Eiji},
  journal={arXiv preprint arXiv:2509.22047},
  year={2025}
}

@article{liu2026gdpo,
  title={GDPO: Group reward-Decoupled Normalization Policy Optimization for Multi-reward RL Optimization},
  author={Liu, Shih-Yang and Dong, Xin and Lu, Ximing and Diao, Shizhe and Belcak, Peter and Liu, Mingjie and Chen, Min-Hung and Yin, Hongxu and Wang, Yu-Chiang Frank and Cheng, Kwang-Ting and others},
  journal={arXiv preprint arXiv:2601.05242},
  year={2026}
}

@article{williams1991function,
  title={Function optimization using connectionist reinforcement learning algorithms},
  author={Williams, Ronald J and Peng, Jing},
  journal={Connection Science},
  volume={3},
  number={3},
  pages={241--268},
  year={1991},
  publisher={Taylor \& Francis}
}

@inproceedings{mnih2016asynchronous,
  title={Asynchronous methods for deep reinforcement learning},
  author={Mnih, Volodymyr and Badia, Adria Puigdomenech and Mirza, Mehdi and Graves, Alex and Lillicrap, Timothy and Harley, Tim and Silver, David and Kavukcuoglu, Koray},
  booktitle={International conference on machine learning},
  pages={1928--1937},
  year={2016},
  organization={PmLR}
}

@inproceedings{haarnoja2018soft,
  title={Soft actor-critic: Off-policy maximum entropy deep reinforcement learning with a stochastic actor},
  author={Haarnoja, Tuomas and Zhou, Aurick and Abbeel, Pieter and Levine, Sergey},
  booktitle={International conference on machine learning},
  pages={1861--1870},
  year={2018},
  organization={Pmlr}
}

@article{liang2025beyond,
  title={Beyond pass@ 1: Self-play with variational problem synthesis sustains rlvr},
  author={Liang, Xiao and Li, Zhongzhi and Gong, Yeyun and Shen, Yelong and Wu, Ying Nian and Guo, Zhijiang and Chen, Weizhu},
  journal={arXiv preprint arXiv:2508.14029},
  year={2025}
}

@article{wei2022chain,
  title={Chain-of-thought prompting elicits reasoning in large language models},
  author={Wei, Jason and Wang, Xuezhi and Schuurmans, Dale and Bosma, Maarten and Xia, Fei and Chi, Ed and Le, Quoc V and Zhou, Denny and others},
  journal={Advances in neural information processing systems},
  volume={35},
  pages={24824--24837},
  year={2022}
}

@article{kojima2022large,
  title={Large language models are zero-shot reasoners},
  author={Kojima, Takeshi and Gu, Shixiang Shane and Reid, Machel and Matsuo, Yutaka and Iwasawa, Yusuke},
  journal={Advances in neural information processing systems},
  volume={35},
  pages={22199--22213},
  year={2022}
}

@inproceedings{
liu2025prorl,
title={Pro{RL}: Prolonged Reinforcement Learning Expands Reasoning Boundaries in Large Language Models},
author={Mingjie Liu and Shizhe Diao and Ximing Lu and Jian Hu and Xin Dong and Yejin Choi and Jan Kautz and Yi Dong},
booktitle={The Thirty-ninth Annual Conference on Neural Information Processing Systems},
year={2025},
url={https://openreview.net/forum?id=YPsJha5HXQ}
}

@article{yang2024qwen2,
  title={Qwen2.5-math technical report: Toward mathematical expert model via self-improvement},
  author={Yang, An and Zhang, Beichen and Hui, Binyuan and Gao, Bofei and Yu, Bowen and Li, Chengpeng and Liu, Dayiheng and Tu, Jianhong and Zhou, Jingren and Lin, Junyang and others},
  journal={arXiv preprint arXiv:2409.12122},
  year={2024}
}

@article{lu2025latent,
  title={Latent Chain-of-Thought? Decoding the Depth-Recurrent Transformer},
  author={Lu, Wenquan and Yang, Yuechuan and Lee, Kyle and Li, Yanshu and Liu, Enqi},
  journal={arXiv preprint arXiv:2507.02199},
  year={2025}
}

@article{dai2026harderl,
      title={Harder Is Better: Boosting Mathematical Reasoning via Difficulty-Aware GRPO and Multi-Aspect Question Reformulation}, 
      author={Yanqi Dai and Yuxiang Ji and Xiao Zhang and Yong Wang and Xiangxiang Chu and Zhiwu Lu},
      journal={arXiv preprint arXiv:2601.20614},
      year={2026},
}

@article{li2025questa,
      title={QuestA: Expanding Reasoning Capacity in LLMs via Question Augmentation}, 
      author={Jiazheng Li and Hongzhou Lin and Hong Lu and Kaiyue Wen and Zaiwen Yang and Jiaxuan Gao and Yi Wu and Jingzhao Zhang},
      journal={arXiv preprint arXiv:2507.13266},
      year={2025},
}

@misc{eval-harness,
  author       = {Gao, Leo and Tow, Jonathan and Abbasi, Baber and Biderman, Stella and Black, Sid and DiPofi, Anthony and Foster, Charles and Golding, Laurence and Hsu, Jeffrey and Le Noac'h, Alain and Li, Haonan and McDonell, Kyle and Muennighoff, Niklas and Ociepa, Chris and Phang, Jason and Reynolds, Laria and Schoelkopf, Hailey and Skowron, Aviya and Sutawika, Lintang and Tang, Eric and Thite, Anish and Wang, Ben and Wang, Kevin and Zou, Andy},
  title        = {The Language Model Evaluation Harness},
  month        = 07,
  year         = 2024,
  publisher    = {Zenodo},
  version      = {v0.4.3},
  doi          = {10.5281/zenodo.12608602},
  url          = {https://zenodo.org/records/12608602}
}

@misc{vonwerra2020trl,
  title   = {{TRL: Transformers Reinforcement Learning}},
  author  = {von Werra, Leandro and Belkada, Younes and Tunstall, Lewis and Beeching, Edward and Thrush, Tristan and Lambert, Nathan and Huang, Shengyi and Rasul, Kashif and Gallouédec, Quentin},
  license = {Apache-2.0},
  url     = {https://github.com/huggingface/trl},
  year    = {2020}
}

@article{sheng2024hybridflow,
  title   = {HybridFlow: A Flexible and Efficient RLHF Framework},
  author  = {Guangming Sheng and Chi Zhang and Zilingfeng Ye and Xibin Wu and Wang Zhang and Ru Zhang and Yanghua Peng and Haibin Lin and Chuan Wu},
  year    = {2024},
  journal = {arXiv preprint arXiv: 2409.19256}
}

@misc{slime_github,
  author       = {Zilin Zhu and Chengxing Xie and Xin Lv and slime Contributors},
  title        = {slime: An LLM post-training framework for RL Scaling},
  year         = {2025},
  howpublished = {\url{https://github.com/THUDM/slime}},
  note         = {GitHub repository. Corresponding author: Xin Lv},
  urldate      = {2025-06-19}
}

@article{li2025treepo,
  title={Treepo: Bridging the gap of policy optimization and efficacy and inference efficiency with heuristic tree-based modeling},
  author={Li, Yizhi and Gu, Qingshui and Wen, Zhoufutu and Li, Ziniu and Xing, Tianshun and Guo, Shuyue and Zheng, Tianyu and Zhou, Xin and Qu, Xingwei and Zhou, Wangchunshu and others},
  journal={arXiv preprint arXiv:2508.17445},
  year={2025}
}

@misc{Kydlicek_MathVerify_2024,
  author  = {Hynek Kydlíček},
  title   = {{Math-Verify}: Math Verification Library},
  version = {0.6.1},
  year    = {2024},
  url     = {https://github.com/huggingface/math-verify},
  license = {Apache-2.0},
}

@article{10.7717/peerj-cs.103,
     title = {SymPy: symbolic computing in Python},
     author = {Meurer, Aaron and Smith, Christopher P. and Paprocki, Mateusz and \v{C}ert\'{i}k, Ond\v{r}ej and Kirpichev, Sergey B. and Rocklin, Matthew and Kumar, AMiT and Ivanov, Sergiu and Moore, Jason K. and Singh, Sartaj and Rathnayake, Thilina and Vig, Sean and Granger, Brian E. and Muller, Richard P. and Bonazzi, Francesco and Gupta, Harsh and Vats, Shivam and Johansson, Fredrik and Pedregosa, Fabian and Curry, Matthew J. and Terrel, Andy R. and Rou\v{c}ka, \v{S}t\v{e}p\'{a}n and Saboo, Ashutosh and Fernando, Isuru and Kulal, Sumith and Cimrman, Robert and Scopatz, Anthony},
     year = 2017,
     month = jan,
     volume = 3,
     pages = {e103},
     journal = {PeerJ Computer Science},
     issn = {2376-5992},
     url = {https://doi.org/10.7717/peerj-cs.103},
     doi = {10.7717/peerj-cs.103}
    }

@article{xie2025interleaved,
  title={Interleaved Reasoning for Large Language Models via Reinforcement Learning},
  author={Xie, Roy and Qiu, David and Gopinath, Deepak and Lin, Dong and Sun, Yanchao and Wang, Chong and Potdar, Saloni and Dhingra, Bhuwan},
  journal={arXiv preprint arXiv:2505.19640},
  year={2025}
}

@inproceedings{
yang2024large,
title={Large Language Models as Optimizers},
author={Chengrun Yang and Xuezhi Wang and Yifeng Lu and Hanxiao Liu and Quoc V Le and Denny Zhou and Xinyun Chen},
booktitle={The Twelfth International Conference on Learning Representations},
year={2024},
url={https://openreview.net/forum?id=Bb4VGOWELI}
}

@article{chen2025dra,
  title={Dra-grpo: Exploring diversity-aware reward adjustment for r1-zero-like training of large language models},
  author={Chen, Xiwen and Zhu, Wenhui and Qiu, Peijie and Dong, Xuanzhao and Wang, Hao and Wu, Haiyu and Li, Huayu and Sotiras, Aristeidis and Wang, Yalin and Razi, Abolfazl},
  journal={arXiv preprint arXiv:2505.09655},
  year={2025}
}

@article{le2026transform,
  title={Transform-Augmented GRPO Improves Pass@ k},
  author={Le, Khiem and Mroueh, Youssef and Nguyen, Phuc and Lin, Chi-Heng and Gao, Shangqian and Hua, Ting and Chawla, Nitesh V},
  journal={arXiv preprint arXiv:2601.22478},
  year={2026}
}

@article{Slow_Thinking_with_LLMs_3_Preview,
  title={STILL-3-1.5B-preview: Enhancing Slow Thinking Abilities of Small Models through Reinforcement Learning
},
  author={RUCAIBox STILL Team},
  url={https://github.com/RUCAIBox/Slow_Thinking_with_LLMs},
  year={2025}
}

@misc{deepscaler2025,
  title={DeepScaleR: Surpassing O1-Preview with a 1.5B Model by Scaling RL},
  author={Michael Luo and Sijun Tan and Justin Wong and Xiaoxiang Shi and William Y. Tang and Manan Roongta and Colin Cai and Jeffrey Luo and Li Erran Li and Raluca Ada Popa and Ion Stoica},
  year={2025},
  howpublished={\href{https://pretty-radio-b75.notion.site/DeepScaleR-Surpassing-O1-Preview-with-a-1-5B-Model-by-Scaling-RL-19681902c1468005bed8ca303013a4e2}{Notion Blog}},
  year={2025}
}

@inproceedings{
    dang2026reinforcement,
    title={Reinforcement Learning for Reasoning in Small {LLM}s: What Works and What Doesn{\textquoteright}t},
    author={Quy-Anh Dang and Chris Ngo},
    booktitle={Logical and Symbolic Reasoning in Language Models @ AAAI 2026},
    year={2026},
    url={https://openreview.net/forum?id=3pWL6Zxc4A}
}

@article{xu2026gac,
  title={GAC: Stabilizing Asynchronous RL Training for LLMs via Gradient Alignment Control},
  author={Xu, Haofeng and Su, Junwei and Tian, Yukun and Diao, Lansong and Qian, Zhengping and Wu, Chuan},
  journal={arXiv preprint arXiv:2603.01501},
  year={2026}
}

@article{li2025repo,
  title={Repo: Replay-enhanced policy optimization},
  author={Li, Siheng and Zhou, Zhanhui and Lam, Wai and Yang, Chao and Lu, Chaochao},
  journal={arXiv preprint arXiv:2506.09340},
  year={2025}
}

@misc{qwen3technicalreport,
      title={Qwen3 Technical Report}, 
      author={Qwen Team},
      year={2025},
      eprint={2505.09388},
      archivePrefix={arXiv},
      primaryClass={cs.CL},
      url={https://arxiv.org/abs/2505.09388}, 
}

@article{qi2025precisionrl,
  title={Defeating the Training-Inference Mismatch via FP16},
  author={Qi, Penghui and Liu, Zichen and Zhou, Xiangxin and Pang, Tianyu and Du, Chao and Lee, Wee Sun and Lin, Min},
  journal={arXiv preprint arXiv:2510.26788},
  year={2025}
}

@online{liu-li-2025-rl-collapse,
  title = {When Speed Kills Stability: Demystifying {RL} Collapse from the Training-Inference Mismatch},
  author = {Liu, Jiacai and Li, Yingru and Fu, Yuqian and Wang, Jiawei and Liu, Qian and Jiang, Zhuo},
  year = {2025},
  month = sep,
  url = {https://richardli.xyz/rl-collapse}
}

@misc{OpenAI_ChatGPT_2026,
  author = {OpenAI},
  title = {ChatGPT},
  year = {2026},
  publisher = {OpenAI},
  address = {https://chat.openai.com/},
  note = {Accessed: 2026-03-20},
  howpublished = {https://chat.openai.com/},
  type = {Large language model}
}

@inproceedings{NEURIPS2025_lu,
 author = {Lu, Wenquan and Zhang, Jiaqi and Van Assel, Hugues and Balestriero, Randall},
 booktitle = {Advances in Neural Information Processing Systems},
 pages = {136057--136089},
 title = {Ditch the Denoiser: Emergence of Noise Robustness in Self-Supervised Learning from Data Curriculum},
 url = {https://proceedings.neurips.cc/paper_files/paper/2025/file/c6a4fc50b10ec1add5086e2f527ff0d3-Paper-Conference.pdf},
 volume = {38},
 year = {2025}
}

@inproceedings{hendrycks2measuring,
  title={Measuring Mathematical Problem Solving With the MATH Dataset},
  author={Hendrycks, Dan and Burns, Collin and Kadavath, Saurav and Arora, Akul and Basart, Steven and Tang, Eric and Song, Dawn and Steinhardt, Jacob},
  booktitle={Thirty-fifth Conference on Neural Information Processing Systems Datasets and Benchmarks Track (Round 2)},
  year={2021}
}

@misc{aime24,
      title={American Invitational Mathematics Examination (AIME) 2024}, 
      author={Zhang, Yifan and Math-AI, Team},
      year={2024},
}

@inproceedings{he2024olympiadbench,
  title={Olympiadbench: A challenging benchmark for promoting agi with olympiad-level bilingual multimodal scientific problems},
  author={He, Chaoqun and Luo, Renjie and Bai, Yuzhuo and Hu, Shengding and Thai, Zhen and Shen, Junhao and Hu, Jinyi and Han, Xu and Huang, Yujie and Zhang, Yuxiang and others},
  booktitle={Proceedings of the 62nd Annual Meeting of the Association for Computational Linguistics (Volume 1: Long Papers)},
  pages={3828--3850},
  year={2024}
}

@article{lewkowycz2022solving,
  title={Solving quantitative reasoning problems with language models},
  author={Lewkowycz, Aitor and Andreassen, Anders and Dohan, David and Dyer, Ethan and Michalewski, Henryk and Ramasesh, Vinay and Slone, Ambrose and Anil, Cem and Schlag, Imanol and Gutman-Solo, Theo and others},
  journal={Advances in neural information processing systems},
  volume={35},
  pages={3843--3857},
  year={2022}
}

@misc{aimo2024validationamc,
  author       = {{AI-MO}},
  title        = {{AIMO Validation AMC Dataset}},
  year         = {2024},
  publisher    = {Hugging Face},
  howpublished = {\url{https://huggingface.co/datasets/AI-MO/aimo-validation-amc}},
  note         = {Accessed: 2026-05-06}
}

@inproceedings{kwon2023efficient,
  title={Efficient Memory Management for Large Language Model Serving with PagedAttention},
  author={Woosuk Kwon and Zhuohan Li and Siyuan Zhuang and Ying Sheng and Lianmin Zheng and Cody Hao Yu and Joseph E. Gonzalez and Hao Zhang and Ion Stoica},
  booktitle={Proceedings of the ACM SIGOPS 29th Symposium on Operating Systems Principles},
  year={2023}
}

@article{schick2023toolformer,
  title={Toolformer: Language models can teach themselves to use tools},
  author={Schick, Timo and Dwivedi-Yu, Jane and Dess{\`\i}, Roberto and Raileanu, Roberta and Lomeli, Maria and Hambro, Eric and Zettlemoyer, Luke and Cancedda, Nicola and Scialom, Thomas},
  journal={Advances in neural information processing systems},
  volume={36},
  pages={68539--68551},
  year={2023}
}

\newpage
\appendix
{\Large \textbf{Outline of the Appendix:}}

\vspace{0.5em}

\begin{itemize}[itemsep=1.0em, topsep=1.0em]
	\item \cref{sec:further_exp}: Further Experiment Results
	\begin{itemize}[itemsep=1.0em, topsep=1.0em]
            \item \cref{sec:std_result}: Evaluation Results with Standard Deviation
            \item \cref{sec:entropy}: Does Prompt Augmentation Increase Token-level Policy Entropy?
            \item \cref{sec:largekl}: Increasing KL Regularization Does Not Match PrAg-PO
        \end{itemize}

        \item \cref{sec:all_prompts}: Prompts and Format Reward Details
        \begin{itemize}[itemsep=1.0em, topsep=1.0em]
            \item \cref{sec:template_stat}: Template Statistics Visualization
            \item \cref{sec:template_qwen}: 13 Templates for Qwen2.5 and Qwen3
            \item \cref{sec:template_deepseek}: 35 Templates for DeepSeek-R1
            \item \cref{sec:format_prac}: Format Reward Best Practices
        \end{itemize}
\end{itemize}
\newpage
\section{Further Experiment Results}
\label{sec:further_exp}

\subsection{Evaluation Results with Standard Deviation}
\label{sec:std_result}
\begin{table}[H]
\setlength\tabcolsep{2pt}
\small
\centering
\caption{Precise evaluation statistics for \cref{tab:main_result_qwen2.5} with standard deviation over the 5 inference rounds.}
\renewcommand{\arraystretch}{1.3}
\begin{tabular}{l c c c c c | c c}
\hline
\textbf{Method} & \textbf{AIME24} & \textbf{AMC}  & \textbf{MATH500} & \textbf{Minerva} & \textbf{Olympiad} & \makecell[c]{\rule{0pt}{2.2ex}\textbf{AVG} \\ \scriptsize benchmark} & \makecell[c]{\rule{0pt}{2.2ex}\textbf{AVG} \\ \scriptsize question} \\\hline

GRPO  & 23.33 $\pm$ 0.00 & 49.88 $\pm$ 1.37 & 75.36 $\pm$ 0.79 & 26.32 $\pm$ 0.49 & 38.28 $\pm$ 0.31 & 42.64 $\pm$ 0.40  & 48.41 $\pm$ 0.42 \\
DAPO  & 23.33 $\pm$ 0.00 & 54.94 $\pm$ 1.37 & 76.92 $\pm$ 0.54 & 25.96 $\pm$ 0.56 & 39.44 $\pm$ 0.32 & 44.12 $\pm$ 0.37 & 49.62 $\pm$ 0.33 \\
\makecell[l]{PrAg-PO\\ (Step 2720)} & 23.33 $\pm$ 0.00 & 53.49 $\pm$ 1.08 & 78.40 $\pm$ 0.45 & 31.25 $\pm$ 0.37 & 39.41 $\pm$ 0.66  & 45.18 $\pm$ 0.19 & 50.92 $\pm$ 0.30 \\
\makecell[l]{PrAg-PO\\ (Step 2480)}  & 20.00 $\pm$ 0.00 & 49.40 $\pm$ 1.20  & 80.40 $\pm$ 0.32 & 29.71 $\pm$ 0.48 & 41.24 $\pm$ 0.59 & 44.15 $\pm$ 0.32  & 51.81 $\pm$ 0.28
\\\hline
\end{tabular}
\label{tab:std_result1}
\end{table}

\begin{table}[H]
\setlength\tabcolsep{2pt}
\small
\centering
\caption{Precise evaluation statistics for \cref{tab:main_result_deepseek} with standard deviation over the 5 inference rounds..}
\renewcommand{\arraystretch}{1.3}
\begin{tabular}{l c c c c c | c c}
\hline
\textbf{Method} & \textbf{AIME24} & \textbf{AMC23}  & \textbf{MATH500} & \textbf{Minerva} & \textbf{Olympiad} & \makecell[c]{\rule{0pt}{2.2ex}\textbf{AVG} \\ \scriptsize benchmark} & \makecell[c]{\rule{0pt}{2.2ex}\textbf{AVG} \\ \scriptsize question} \\\hline

GRPO  & 30.00 $\pm$ 0.00 & 74.00 $\pm$ 1.37 & 85.48 $\pm$ 0.18 & 28.24 $\pm$ 0.71 & 46.49 $\pm$ 0.12 & 52.84 $\pm$ 0.28  & 56.47 $\pm$ 0.14 \\
DAPO  & 33.33 $\pm$ 0.00 & 77.50 $\pm$ 1.77 & 84.24 $\pm$ 0.97 & 26.91 $\pm$ 1.57 & 50.04 $\pm$ 0.91 & 54.41 $\pm$ 0.34 & 57.56 $\pm$ 0.46 \\
\makecell[l]{PrAg-PO\\ (Step 1100)} & 42.00 $\pm$ 2.98 & 78.50 $\pm$ 3.35 & 87.48 $\pm$ 0.73 & 29.34 $\pm$ 0.92 & 53.81 $\pm$ 0.64  & 58.23 $\pm$ 1.10 & 60.94 $\pm$ 0.18 \\
\makecell[l]{PrAg-PO\\ (Step 1160)}  & 33.33 $\pm$ 0.00 & 77.00 $\pm$ 2.74 & 87.24 $\pm$ 1.05 & 32.65 $\pm$ 0.16 & 56.18 $\pm$ 0.83  & 57.28 $\pm$ 0.84 & 62.29 $\pm$ 0.63
\\\hline
\end{tabular}
\label{tab:std_result2}
\end{table}

\begin{table}[H]
\setlength\tabcolsep{2pt}
\small
\centering
\caption{Precise evaluation statistics for \cref{tab:main_result_qwen3} with standard deviation over the 5 inference rounds..}
\renewcommand{\arraystretch}{1.3}
\begin{tabular}{l c c c c c | c c}
\hline
\textbf{Method} & \textbf{AIME24} & \textbf{AMC23}  & \textbf{MATH500} & \textbf{Minerva} & \textbf{Olympiad} & \makecell[c]{\rule{0pt}{2.2ex}\textbf{AVG} \\ \scriptsize benchmark} & \makecell[c]{\rule{0pt}{2.2ex}\textbf{AVG} \\ \scriptsize question} \\\hline

GRPO  & 16.67 $\pm$ 0.00 & 70.00 $\pm$ 0.00 & 78.72 $\pm$ 0.87 & 33.16 $\pm$ 0.55 & 45.13 $\pm$ 0.37 & 48.73 $\pm$ 0.13  & 54.15 $\pm$ 0.29 \\
DAPO  & 26.67 $\pm$ 0.00 & 70.00 $\pm$ 0.00 & 84.88 $\pm$ 0.39 & 32.79 $\pm$ 0.66 & 55.05 $\pm$ 0.13 & 53.88 $\pm$ 0.16 & 60.73 $\pm$ 0.20 \\
\makecell[l]{PrAg-PO\\ (Step 1520)} & 33.33 $\pm$ 0.00 & 75.00 $\pm$ 0.00 & 85.12 $\pm$ 0.23 & 33.60 $\pm$ 0.56 & 53.87 $\pm$ 0.13  & 56.18 $\pm$ 0.12 & 60.69 $\pm$ 0.13 \\
\makecell[l]{PrAg-PO\\ (Step 720)}  & 23.33 $\pm$ 0.00 & 76.50 $\pm$ 1.37  & 87.04 $\pm$ 0.43 & 33.53 $\pm$ 1.20 & 55.67 $\pm$ 0.44 & 55.22 $\pm$ 0.13  & 61.95 $\pm$ 0.25
\\\hline
\end{tabular}
\label{tab:std_result3}
\end{table}

\subsection{Does Prompt Augmentation Increase Token-level Policy Entropy?}
\label{sec:entropy}
\begin{figure}[t]
    \centering
    \includegraphics[width=0.6\linewidth]{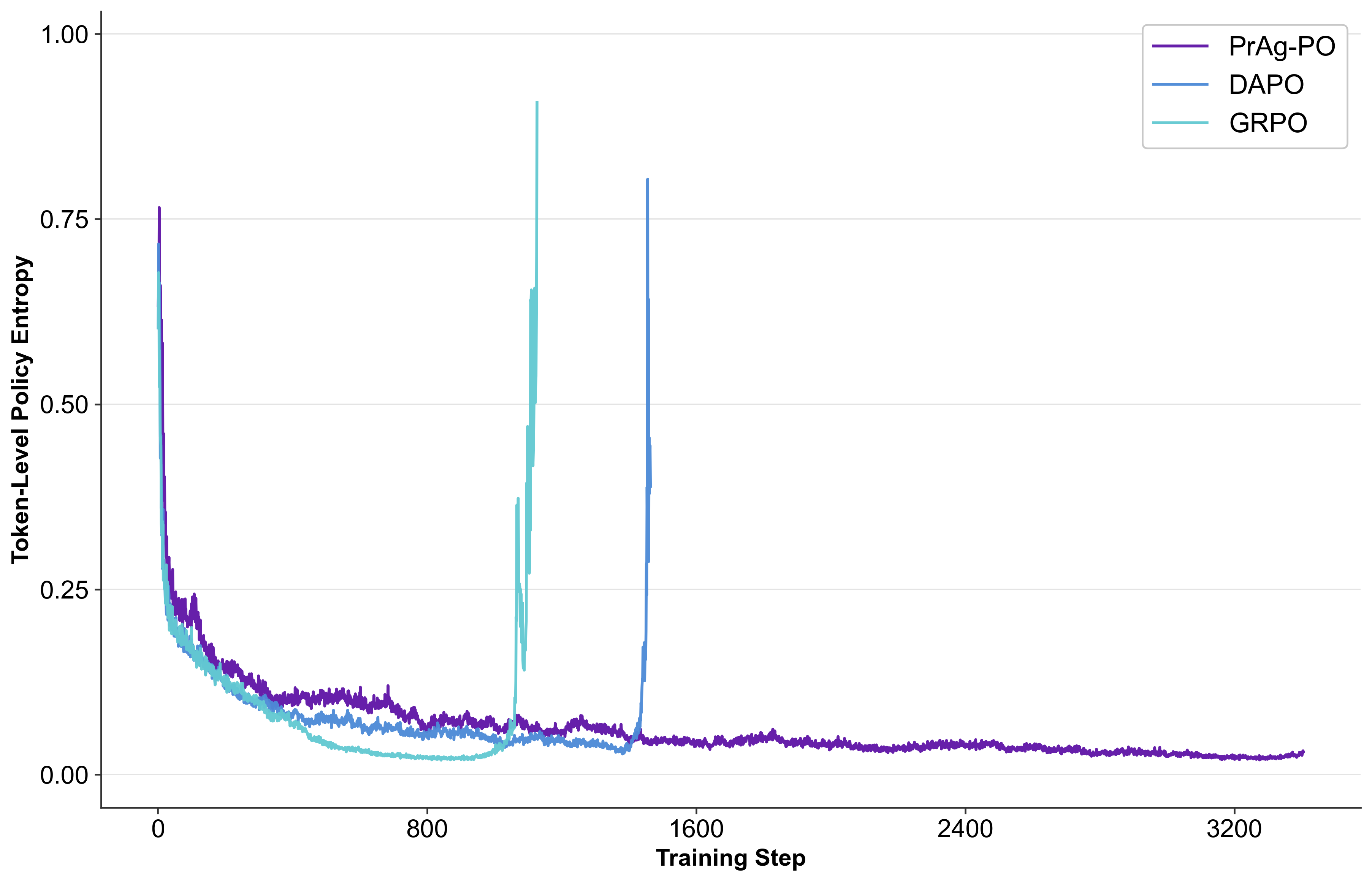}
    \caption{Token-level policy entropy during training of different methods on Qwen2.5-Math-1.5B.}
    \label{fig:qwen2_5_entropy}
    \vspace{-10pt}
\end{figure}

\begin{figure}[H]
    \centering
    \includegraphics[width=0.6\linewidth]{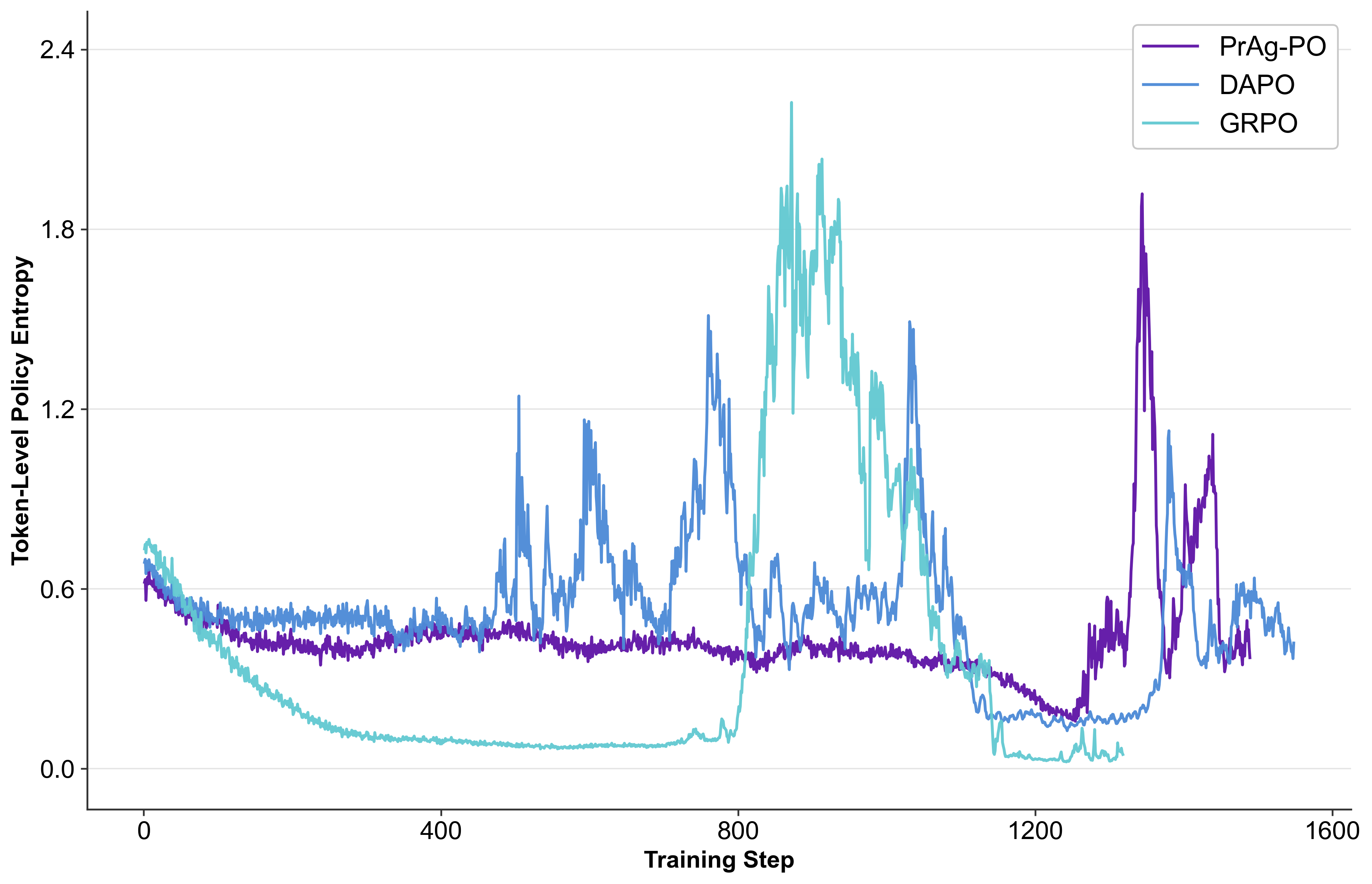}
    \caption{Token-level policy entropy during training of different methods on DeepSeek-R1-Distill-Qwen-1.5B}
    \label{fig:deepseek_entropy}
    \vspace{-10pt}
\end{figure}

\begin{figure}[H]
    \centering
    \includegraphics[width=0.6\linewidth]{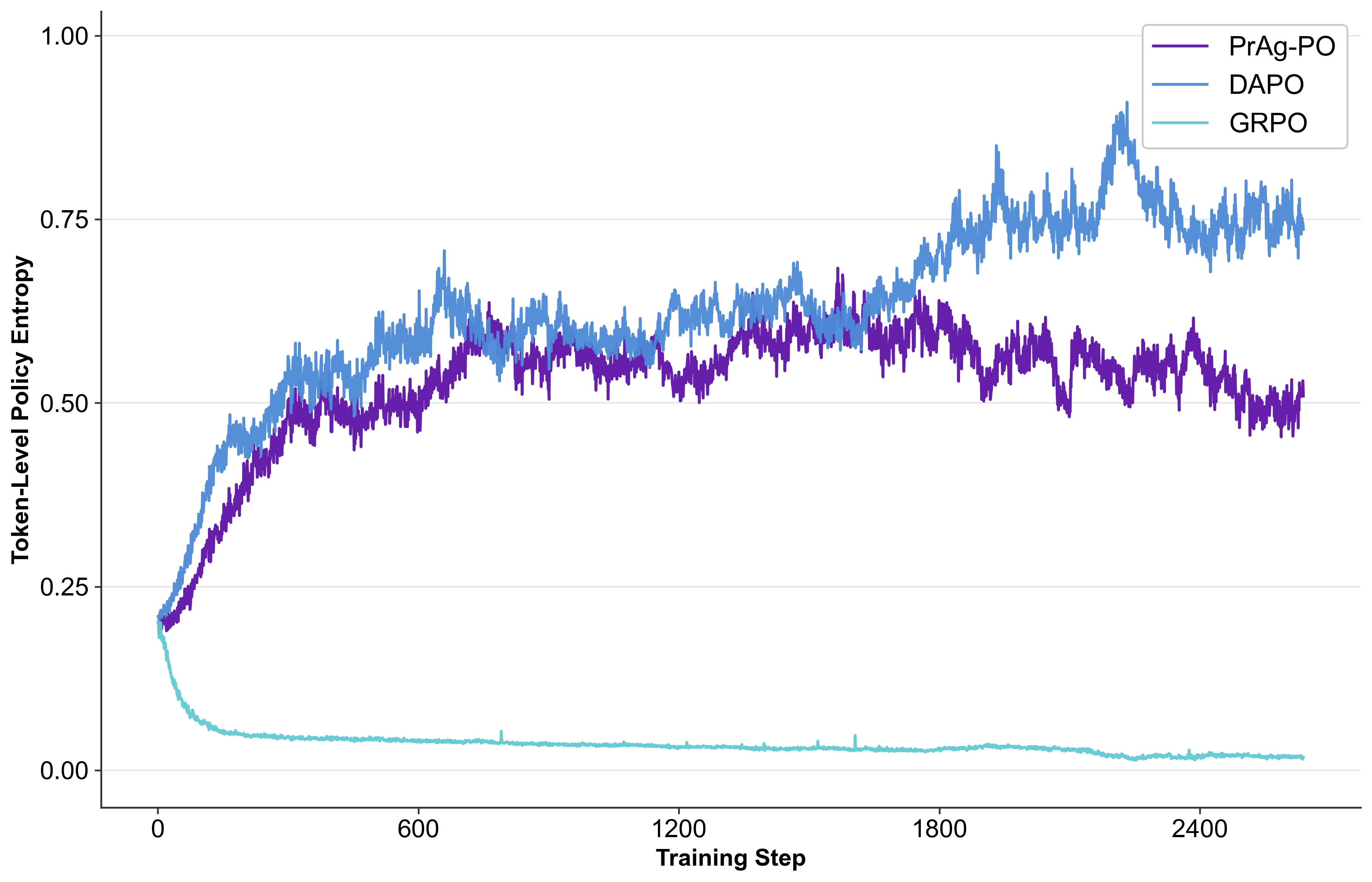}
    \caption{Token-level policy entropy during training of different methods on Qwen3-1.7B}
    \label{fig:qwen3_entropy}
\end{figure}

Given that prompt augmentation elicits diverse reasoning paths and output formats, a natural question is whether it also regularizes training by increasing token-level policy entropy. We find that the answer is nuanced: \textbf{As shown in \cref{fig:qwen2_5_entropy}, \cref{fig:deepseek_entropy}, and \cref{fig:qwen3_entropy}, prompt augmentation helps stabilizing policy entropy instead of merely increasing it}.

For a large language model parameterized by $\theta$, the policy $\pi_\theta$ defines a conditional distribution over the next token given the input question $q$ and the previously generated tokens $o_{<t}$. The token-level policy entropy at decoding step $t$ is defined as:
$$
\mathcal{H}_{t}(\pi_\theta)
= - \sum_{v \in \mathcal{V}} \pi_\theta(v \mid q, o_{<t}) \log \pi_\theta(v \mid q, o_{<t}),
$$
where $\mathcal{V}$ denotes the vocabulary. In practice, token-level policy entropy  aggregated over generated response tokens is computed as
$$
\mathcal{H}_{\mathrm{agg}}(\pi_\theta)
=
\frac{1}{\sum_{i=1}^{B} |o_i|}
\sum_{i=1}^{B} \sum_{t=1}^{|o_i|}
\mathcal{H}_{i,t}(\pi_\theta),
$$

where $o_i$ denotes the $i$-th generated response in a batch $B$. Higher entropy corresponds to more stochastic and exploratory generation behavior. However, while token-level entropy measures \textbf{local uncertainty}, true diversity is inherently a sequence-level property. In principle, sequence-level entropy provides a more faithful characterization of this diversity, but it is intractable to compute for autoregressive models.

As illustrated in \cref{fig:qwen2_5_entropy}, \cref{fig:deepseek_entropy}, and \cref{fig:qwen3_entropy}, training collapse is often accompanied by sharp entropy spikes and unstable fluctuations. While the token-level policy entropy of PrAg-PO (purple) is not consistently higher than that of DAPO, its entropy curve is notably smoother and exhibits a delayed onset of large fluctuations. This suggests that prompt augmentation stabilizes token-level policy entropy during training, rather than simply increasing it, while still promoting diversity through variation in reasoning paths and output formats.

\subsection{Increasing KL Regularization Does Not Match PrAg-PO}
\label{sec:largekl}
\begin{figure}[H]
    \centering
    \includegraphics[width=0.6\linewidth]{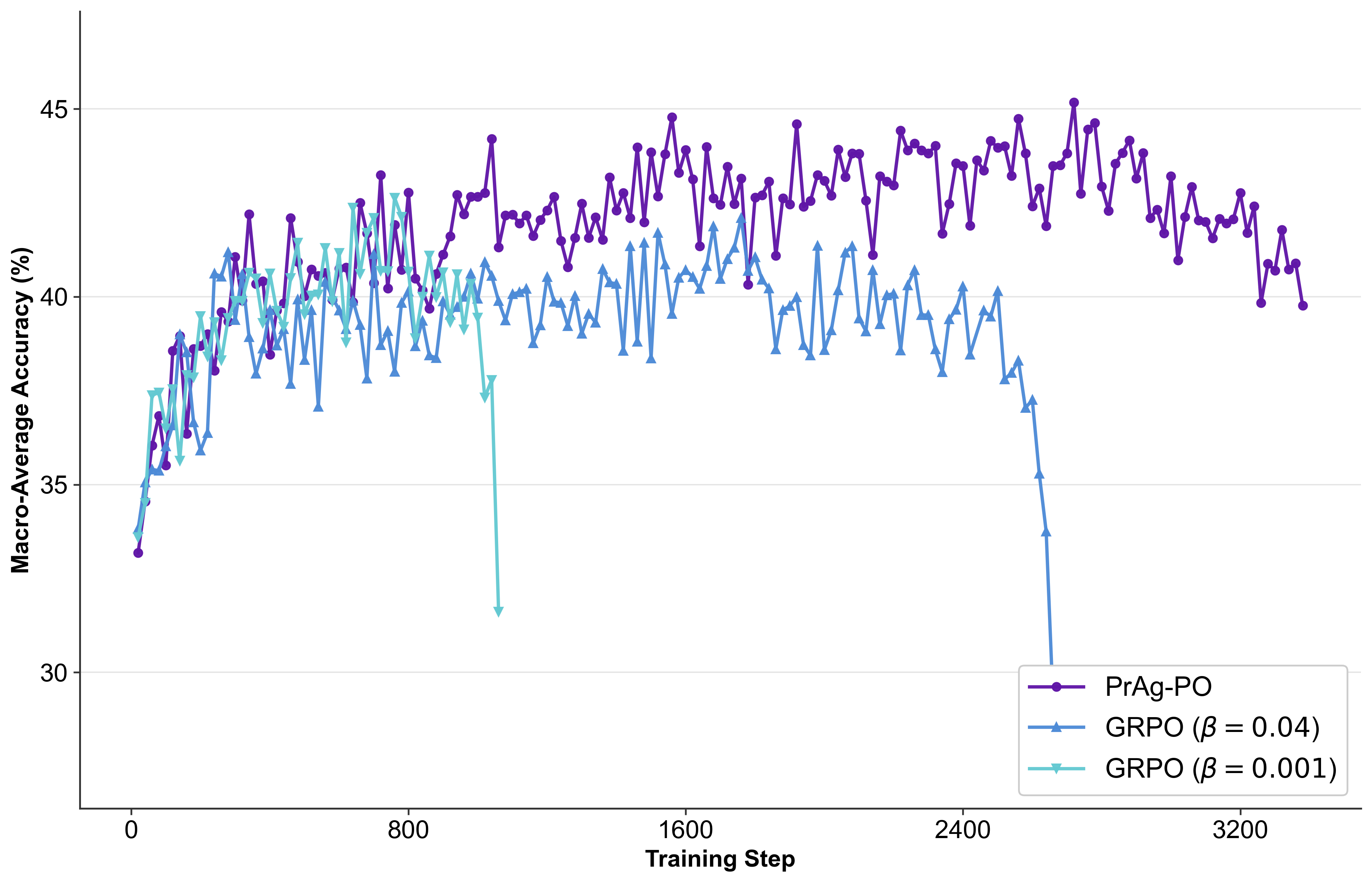}
    \caption{Evaluation accuracy over training steps with varying KL regularization coefficient. Increasing the KL coefficient in GRPO (darker blue) delays collapse but yields substantially lower accuracy.}
    \label{fig:large_kl}
\end{figure}

Strong KL regularization constrains the model within narrower trust regions and has the potential to stabilize training, mitigate entropy collapse, and enable long-horizon optimization. Accordingly, we conduct GRPO training with a substantially larger KL coefficient ($\beta = 0.04$, as originally used in TRL) over an extended training horizon. As shown in \cref{fig:large_kl}, although strong KL regularization effectively delays training collapse, it yields significantly lower performance than PrAg-PO. This is because strong KL constraints restrict exploratory behavior and keep the policy overly close to the base model. These results further highlight the effectiveness of PrAg-PO.

\newpage
\section{Prompts and Format Rewards Details}\label{sec:all_prompts}
\subsection{Template Statistics Visualization}
\label{sec:template_stat}
\begin{figure}[H]
    \centering
    \includegraphics[width=\linewidth]{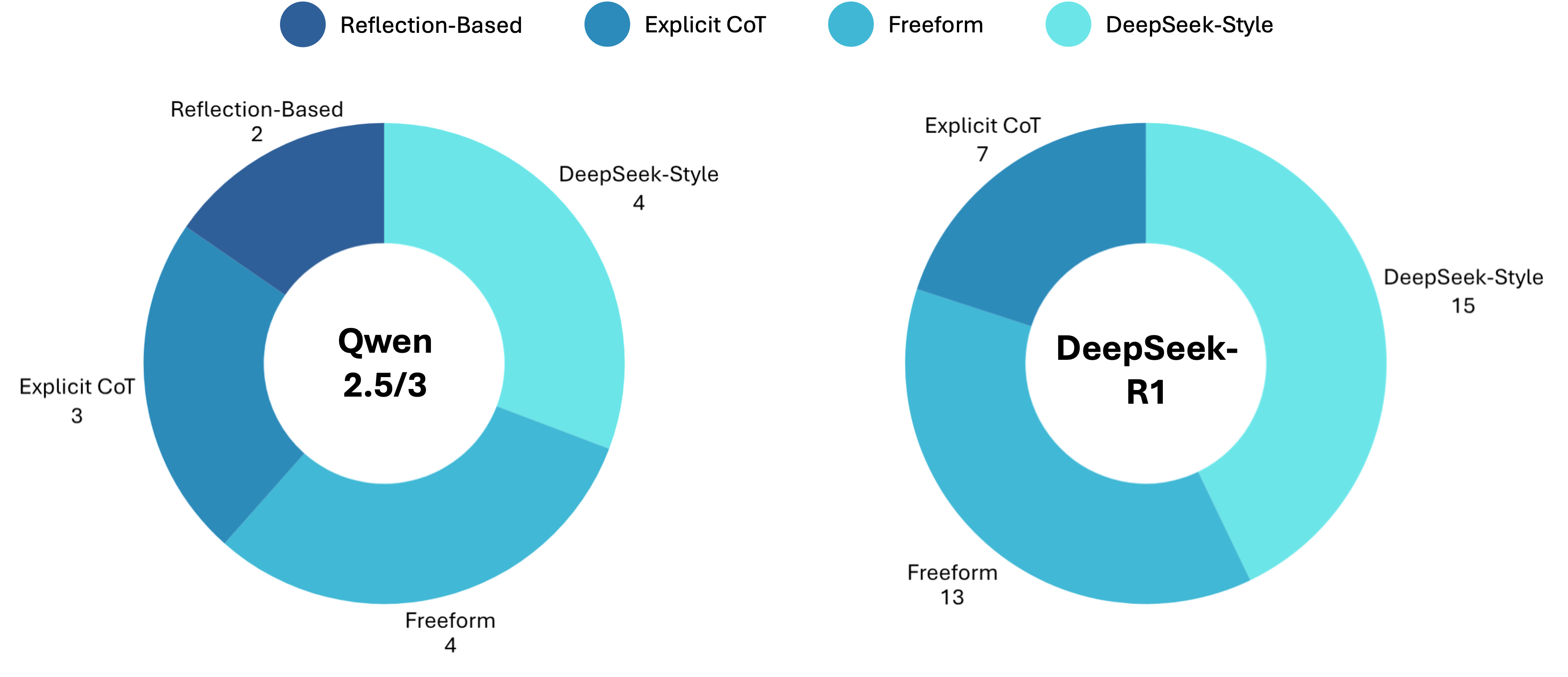}
    \caption{Template category distribution for Qwen2.5-Math-1.5B and Qwen3-1.7B (left), and DeepSeek-R1-Distill-Qwen-1.5B (right). Reflective templates are removed for DeepSeek-R1 because we observe that they slightly degrade training accuracy. We observe that DeepSeek-R1-Distill-Qwen-1.5B already exhibits frequent reflection and self-doubt behavior, and training with reflective templates can make this behavior excessive.}
    \label{fig:template_donut}
\end{figure}
\subsection{13 Templates for Qwen2.5 and Qwen3}
\label{sec:template_qwen}
\lstset{
  basicstyle=\ttfamily\footnotesize,
  breaklines=true,
  backgroundcolor=\color{gray!10},
  frame=single,
  columns=fullflexible,
  captionpos=b,
  keywordstyle=\color{blue},
  commentstyle=\color{gray}\itshape,
  showstringspaces=false
}
\begin{tabularx}{\textwidth}{ | m{6cm} | m{1.8cm} | m{5cm} | }
\caption{13 initial prompt templates used to train Qwen2.5-Math-1.5B and Qwen3-1.7B models, their associated categorizations, and format reward functions.}
\label{tab:all_prompts} \\
  \hline
  \textbf{Prompt Template} & \textbf{Category} & \textbf{Format Reward Function}
    \\\hline
  \noindent\parbox[b]{\hsize}{\textless\textbar{}im\_start\textbar{}\textgreater system\\ Please reason step by step, and put your final answer within \texttt{\textbackslash boxed\{\}}.\textless\textbar{}im\_end\textbar{}\textgreater\\\textless\textbar{}im\_start\textbar{}\textgreater user\\\{question\}\textless\textbar{}im\_end\textbar{}\textgreater\\\textless\textbar{}im\_start\textbar{}\textgreater assistant\\Let's think step by step.} & Explicit CoT & 
  \begin{lstlisting}[language=Python]
^^Jdef format_reward(completion):^^J \ \ return 1.0
\end{lstlisting}\\
  \hline
\noindent\parbox[b]{\hsize}{\textless\textbar{}im\_start\textbar{}\textgreater system\\ You are a helpful assistant.\textless\textbar{}im\_end\textbar{}\textgreater\\\textless\textbar{}im\_start\textbar{}\textgreater user\\\{question\}\\ Please reason step by step, and put your final answer within \texttt{\textbackslash boxed\{\}}.\textless\textbar{}im\_end\textbar{}\textgreater\\\textless\textbar{}im\_start\textbar{}\textgreater assistant} & Freeform & 
\begin{lstlisting}[language=Python]
^^Jdef format_reward(completion):^^J \ \ return 1.0
\end{lstlisting}
\\\hline
\noindent\parbox[b]{\hsize}{
\textless\textbar{}im\_start\textbar{}\textgreater system\\ Please reason step by step, and put your final answer within \texttt{\textbackslash boxed\{\}}.\textless\textbar{}im\_end\textbar{}\textgreater\\\textless\textbar{}im\_start\textbar{}\textgreater user\\\{question\}\textless\textbar{}im\_end\textbar{}\textgreater\\\textless\textbar{}im\_start\textbar{}\textgreater assistant
}
& Freeform & \begin{lstlisting}[language=Python]
^^Jdef format_reward(completion):^^J \ \ return 1.0
\end{lstlisting}\\\hline
\noindent\parbox[b]{\hsize}{
\textless\textbar{}im\_start\textbar{}\textgreater system\\ You are a helpful AI Assistant that provides well-reasoned and detailed responses. You first think about the reasoning process as an internal monologue and then provide the user with the answer. Respond in the following format: \textless think\textgreater\\ ...\\ \textless/think\textgreater\\ \textless answer\textgreater \\...\\ \textless/answer\textgreater \\ Inside the \textless answer\textgreater...\textless/answer\textgreater block, the final answer must be enclosed in \texttt{\textbackslash boxed\{\}}.\textless\textbar{}im\_end\textbar{}\textgreater\\\textless\textbar{}im\_start\textbar{}\textgreater user\\\{question\}\textless\textbar{}im\_end\textbar{}\textgreater\\\textless\textbar{}im\_start\textbar{}\textgreater assistant
} & DeepSeek-Style & \begin{lstlisting}[language=Python]
^^Jdef format_reward(completion):^^J \ \ count = 0.0^^J \ \ if completion.count("<think>\\n") == 1:^^J \ \ \ \ count += 0.25^^J \ \ if completion.count("\\n</think>\\n") == 1:^^J \ \ \ \ count += 0.25^^J \ \ if completion.count("\\n<answer>\\n") == 1:^^J \ \ \ \ count += 0.25^^J \ \ if completion.count("\\n</answer>") == 1:^^J \ \ \ \ count += 0.25^^J \ \ return count
\end{lstlisting}
\\\hline 
\noindent\parbox[b]{\hsize}{
\textless\textbar{}im\_start\textbar{}\textgreater system\\ You are a helpful AI Assistant that provides well-reasoned and detailed responses. You first think about the reasoning process as an internal monologue and then provide the user with the answer. Respond in the following format: \textless think\textgreater\\ ...\\ \textless/think\textgreater\\ \textless answer\textgreater \\...\\ \textless/answer\textgreater \\ Inside the \textless answer\textgreater...\textless/answer\textgreater block, the final answer must be enclosed in \texttt{\textbackslash boxed\{\}}.\textless\textbar{}im\_end\textbar{}\textgreater\\\textless\textbar{}im\_start\textbar{}\textgreater user\\\{question\}\textless\textbar{}im\_end\textbar{}\textgreater\\\textless\textbar{}im\_start\textbar{}\textgreater assistant\\\textless/think\textgreater
} & DeepSeek-Style (Teacher-Forced) & \begin{lstlisting}[language=Python]
^^Jdef format_reward(completion):^^J \ \ count = 0.0^^J \ \ if completion.count("\\n</think>\\n") == 1:^^J \ \ \ \ count += 1/3^^J \ \ if completion.count("\\n<answer>\\n") == 1:^^J \ \ \ \ count += 1/3^^J \ \ if completion.count("\\n</answer>") == 1:^^J \ \ \ \ count += 1/3^^J \ \ return count
\end{lstlisting}
\\\hline
\noindent\parbox[b]{\hsize}{
\textless\textbar{}im\_start\textbar{}\textgreater system\\ A conversation between User and Assistant. The User asks a question, and the Assistant solves it. The Assistant first thinks about the reasoning process in the mind and then provides the User with the answer. The reasoning process is enclosed within \textless think\textgreater \ \textless/think\textgreater \ and answer is enclosed within \textless answer\textgreater \ \textless/answer\textgreater \ tags, respectively, i.e., \textless think\textgreater \ reasoning process here \textless/think\textgreater \ \textless answer\textgreater \ answer here \textless/answer\textgreater. Inside the \textless answer\textgreater...\textless/answer\textgreater \ block, the final answer must be enclosed in \texttt{\textbackslash boxed\{\}}.\textless\textbar{}im\_end\textbar{}\textgreater\\\textless\textbar{}im\_start\textbar{}\textgreater user\\\{question\}\textless\textbar{}im\_end\textbar{}\textgreater\\\textless\textbar{}im\_start\textbar{}\textgreater assistant
} & DeepSeek-Style & \begin{lstlisting}[language=Python]
^^Jdef format_reward(completion):^^J \ \ count = 0.0^^J \ \ if completion.count("<think>") == 1:^^J \ \ \ \ count += 0.25^^J \ \ if completion.count("</think>") == 1:^^J \ \ \ \ count += 0.25^^J \ \ if completion.count("<answer>") == 1:^^J \ \ \ \ count += 0.25^^J \ \ if completion.count("</answer>") == 1:^^J \ \ \ \ count += 0.25^^J \ \ return count
\end{lstlisting}
\\\hline
\noindent\parbox[b]{\hsize}{
\textless\textbar{}im\_start\textbar{}\textgreater system\\ A conversation between User and Assistant. The User asks a question, and the Assistant solves it. The Assistant first thinks about the reasoning process in the mind and then provides the User with the answer. The reasoning process is enclosed within \textless think\textgreater \ \textless/think\textgreater \ and answer is enclosed within \textless answer\textgreater \ \textless/answer\textgreater \ tags, respectively, i.e., \textless think\textgreater \ reasoning process here \textless/think\textgreater \ \textless answer\textgreater \ answer here \textless/answer\textgreater. Inside the \textless answer\textgreater...\textless/answer\textgreater \ block, the final answer must be enclosed in \texttt{\textbackslash boxed\{\}}.\textless\textbar{}im\_end\textbar{}\textgreater\\\textless\textbar{}im\_start\textbar{}\textgreater user\\\{question\}\textless\textbar{}im\_end\textbar{}\textgreater\\\textless\textbar{}im\_start\textbar{}\textgreater assistant\\ \textless think\textgreater
} & DeepSeek-Style (Teacher-Forced) & \begin{lstlisting}[language=Python]
^^Jdef format_reward(completion):^^J \ \ count = 0.0^^J  \ \ if completion.count("</think>") == 1:^^J \ \ \ \ count += 1/3^^J \ \ if completion.count("<answer>") == 1:^^J \ \ \ \ count += 1/3^^J \ \ if completion.count("</answer>") == 1:^^J \ \ \ \ count += 1/3^^J \ \ return count
\end{lstlisting}
\\\hline
\noindent\parbox[b]{\hsize}{
\textless\textbar{}im\_start\textbar{}\textgreater system\\ You are an intelligent assistant who helps with user questions. Provide a rigorous, step-by-step derivation of the solution. The final answer must be clearly indicated within \texttt{\textbackslash boxed\{\}}.\textless\textbar{}im\_end\textbar{}\textgreater\\\textless\textbar{}im\_start\textbar{}\textgreater user\\\{question\}\textless\textbar{}im\_end\textbar{}\textgreater\\\textless\textbar{}im\_start\textbar{}\textgreater assistant
}
& Freeform & 
\begin{lstlisting}[language=Python]
^^Jdef format_reward(completion):^^J \ \ return 1.0
\end{lstlisting}
\\\hline
\noindent\parbox[b]{\hsize}{
\textless\textbar{}im\_start\textbar{}\textgreater system\\ Solve the following math challenge. Explain your approach step-by-step\\The answer should end with: The final answer is: \texttt{\textbackslash boxed\{answer\}}\\where [answer] is just the final number or expression that solves the problem.\textless\textbar{}im\_end\textbar{}\textgreater\\\textless\textbar{}im\_start\textbar{}\textgreater user\\\{question\}\textless\textbar{}im\_end\textbar{}\textgreater\\\textless\textbar{}im\_start\textbar{}\textgreater assistant\\Let's think step by step
}
& Explicit CoT & 
\begin{lstlisting}[language=Python]
^^Jdef format_reward(completion):^^J \ \ if completion.count("The final answer is:") == 1: ^^J \ \ \ \ return 1.0^^J  \ \ else:^^J \ \ \ \  return 0.0
\end{lstlisting}
\\\hline
\noindent\parbox[b]{\hsize}{\textless\textbar{}im\_start\textbar{}\textgreater system\\ Analyze and solve the math task.\textless\textbar{}im\_end\textbar{}\textgreater\\\textless\textbar{}im\_start\textbar{}\textgreater user\\\{question\}\\ End the answer with:\\The final answer is: \texttt{\textbackslash boxed\{answer\}} where [answer] is just the final number or expression that solves the problem.\textless\textbar{}im\_end\textbar{}\textgreater\\\textless\textbar{}im\_start\textbar{}\textgreater assistant} & Freeform & 
\begin{lstlisting}[language=Python]
^^Jdef format_reward(completion):^^J \ \ if completion.count("The final answer is:") == 1: ^^J \ \ \ \ return 1.0^^J  \ \ else:^^J \ \ \ \  return 0.0
\end{lstlisting}
\\\hline
\noindent\parbox[b]{\hsize}{
\textless\textbar{}im\_start\textbar{}\textgreater system\\ Solve the following math problem\\Show each step of your solution\\Put the final answer within \texttt{\textbackslash boxed\{answer\}}\\where [answer] is just the final number or expression that solves the problem.\textless\textbar{}im\_end\textbar{}\textgreater\\\textless\textbar{}im\_start\textbar{}\textgreater user\\\{question\}\textless\textbar{}im\_end\textbar{}\textgreater\\\textless\textbar{}im\_start\textbar{}\textgreater assistant\\Let's think step by step
}
& Explicit CoT & 
\begin{lstlisting}[language=Python]
^^Jdef format_reward(completion):^^J \ \ return 1.0
\end{lstlisting}
\\\hline
\noindent\parbox[b]{\hsize}{
\textless\textbar{}im\_start\textbar{}\textgreater system\\ You are a helpful assistant that solves math problems. Always write out your reasoning to produce a solution, then check whether the solution is correct, fix it if it is wrong, and finally give the final answer. Respond in exactly the following format: \textless solution\textgreater\\reasoning and solution\\\textless/solution\textgreater\\\textless check\textgreater\\Let's verify step by step ...\\\textless/check\textgreater\\Put your final answer within \texttt{\textbackslash boxed\{\}}.\textless\textbar{}im\_end\textbar{}\textgreater\\\textless\textbar{}im\_start\textbar{}\textgreater user\\\{question\}\textless\textbar{}im\_end\textbar{}\textgreater\\\textless\textbar{}im\_start\textbar{}\textgreater assistant
}
& Reflection-Based & 
\begin{lstlisting}[language=Python]
^^Jdef format_reward(completion):^^J \ \ count = 0.0^^J \ \ if completion.count("<solution>\\n") == 1:^^J \ \ \ \ count += 0.25^^J \ \ if completion.count("\\n</solution>\\n") == 1:^^J \ \ \ \ count += 0.25^^J \ \ if completion.count("\\n<check>\\n Let's verify step by step") == 1:^^J \ \ \ \ count += 0.25^^J \ \ if completion.count("</answer>") == 1:^^J \ \ \ \ count += 0.25^^J \ \ return count
\end{lstlisting}
\\\hline
\noindent\parbox[b]{\hsize}{
\textless\textbar{}im\_start\textbar{}\textgreater system\\ You are a helpful assistant that solves math problems. Always write out your reasoning to produce a solution, then check whether the solution is correct, fix it if it is wrong, and finally give the final answer. Respond in exactly the following format: \textless solution\textgreater\\reasoning and solution\\\textless/solution\textgreater\\\textless check\textgreater\\Let's verify step by step ...\\\textless/check\textgreater\\Put your final answer within \texttt{\textbackslash boxed\{\}}.\textless\textbar{}im\_end\textbar{}\textgreater\\\textless\textbar{}im\_start\textbar{}\textgreater user\\\{question\}\textless\textbar{}im\_end\textbar{}\textgreater\\\textless\textbar{}im\_start\textbar{}\textgreater assistant\\ \textless solution\textgreater
}
& Reflection-Based (Teacher-Forced) & 
\begin{lstlisting}[language=Python]
^^Jdef format_reward(completion):^^J \ \ count = 0.0^^J \ \ if completion.count("\\n</solution>\\n") == 1:^^J \ \ \ \ count += 1/3^^J \ \ if completion.count("\\n<check>\\n Let's verify step by step") == 1:^^J \ \ \ \ count += 1/3^^J \ \ if completion.count("\\n</check>") == 1:^^J \ \ \ \ count += 1/3^^J \ \ return count
\end{lstlisting}
\\\hline
\end{tabularx}


\newpage
\subsection{35 Templates for DeepSeek-R1}
\label{sec:template_deepseek}
\begin{tabularx}{\textwidth}{ | m{6cm} | m{1.8cm} | m{5cm} | }
\caption{35 prompt templates used to train the DeepSeek-R1-Distill-Qwen-1.5B model, their associated categorizations, and format reward functions. Note that DeepSeek-R1 use a different default template than Qwen, so we first map the initial Qwen templates to DeepSeek style (e.g., `assistant' to `\textless\textbar{}Assistant\textbar{}\textgreater'). The first 11 templates come from the initial template set, and the next 23 templates are generated by a proprietary LLM. The last template is the default template of DeepSeek-R1-Distill-Qwen-1.5B. We removed reflective templates for this training because we found that they slightly degraded training accuracy. We observe that DeepSeek-R1-Distill-Qwen-1.5B already exhibits frequent reflection and self-doubt behavior, and training with reflective templates can make this behavior excessive. We add line breaks for presentation clarity; please refer to our codebase for the exact template strings.}. 
\label{tab:all_prompts} \\
  \hline
  \textbf{Prompt Template} & \textbf{Category} & \textbf{Format Reward Function}
  \\\hline
\noindent\parbox[b]{\hsize}{
\textless bos\textgreater{}Please reason step by step, and put your final answer within \texttt{\textbackslash boxed\{\}}.\\
\textless\textbar{}User\textbar{}\textgreater\{question\}\\
\textless\textbar{}Assistant\textbar{}\textgreater Let's think step by step.
}
& Explicit CoT &
\begin{lstlisting}[language=Python]
^^Jdef format_reward(completion):^^J \ \ return 1.0
\end{lstlisting}
\\\hline
\noindent\parbox[b]{\hsize}{
\textless bos\textgreater{}You are a helpful assistant.\\
\textless\textbar{}User\textbar{}\textgreater\{question\}\\
Please reason step by step, and put your final answer within \texttt{\textbackslash boxed\{\}}.\\
\textless\textbar{}Assistant\textbar{}\textgreater
}
& Freeform &
\begin{lstlisting}[language=Python]
^^Jdef format_reward(completion):^^J \ \ return 1.0
\end{lstlisting}
\\\hline
\noindent\parbox[b]{\hsize}{
\textless bos\textgreater{}Please reason step by step, and put your final answer within \texttt{\textbackslash boxed\{\}}.\\
\textless\textbar{}User\textbar{}\textgreater\{question\}\\
\textless\textbar{}Assistant\textbar{}\textgreater
}
& Freeform &
\begin{lstlisting}[language=Python]
^^Jdef format_reward(completion):^^J \ \ return 1.0
\end{lstlisting}
\\\hline
\noindent\parbox[b]{\hsize}{
\textless bos\textgreater{}You are a helpful AI Assistant that provides well-reasoned and detailed responses. You first think about the reasoning process as an internal monologue and then provide the user with the answer. Respond in the following format: \textless think\textgreater\\
...\\
\textless/think\textgreater\\
\textless answer\textgreater\\
...\\
\textless/answer\textgreater\\
Inside the \textless answer\textgreater...\textless/answer\textgreater block, the final answer must be enclosed in \texttt{\textbackslash boxed\{\}}.\\
\textless\textbar{}User\textbar{}\textgreater\{question\}\\
\textless\textbar{}Assistant\textbar{}\textgreater
}
& DeepSeek-Style &
\begin{lstlisting}[language=Python]
^^Jdef format_reward(completion):^^J \ \ count = 0.0^^J \ \ if completion.count("<think>\\n") == 1:^^J \ \ \ \ count += 0.25^^J \ \ if completion.count("\\n</think>\\n") == 1:^^J \ \ \ \ count += 0.25^^J \ \ if completion.count("\\n<answer>\\n") == 1:^^J \ \ \ \ count += 0.25^^J \ \ if completion.count("\\n</answer>") == 1:^^J \ \ \ \ count += 0.25^^J \ \ return count
\end{lstlisting}
\\\hline
\noindent\parbox[b]{\hsize}{
\textless bos\textgreater{}You are a helpful AI Assistant that provides well-reasoned and detailed responses. You first think about the reasoning process as an internal monologue and then provide the user with the answer. Respond in the following format: \textless think\textgreater\\
...\\
\textless/think\textgreater\\
\textless answer\textgreater\\
...\\
\textless/answer\textgreater\\
Inside the \textless answer\textgreater...\textless/answer\textgreater block, the final answer must be enclosed in \texttt{\textbackslash boxed\{\}}.\\
\textless\textbar{}User\textbar{}\textgreater\{question\}\\
\textless\textbar{}Assistant\textbar{}\textgreater\textless think\textgreater
}
& DeepSeek-Style (Teacher-Forced) &
\begin{lstlisting}[language=Python]
^^Jdef format_reward(completion):^^J \ \ count = 0.0^^J \ \ if completion.count("\\n</think>\\n") == 1:^^J \ \ \ \ count += 1/3^^J \ \ if completion.count("\\n<answer>\\n") == 1:^^J \ \ \ \ count += 1/3^^J \ \ if completion.count("\\n</answer>") == 1:^^J \ \ \ \ count += 1/3^^J \ \ return count
\end{lstlisting}
\\\hline
\noindent\parbox[b]{\hsize}{
\textless bos\textgreater{}A conversation between User and Assistant. The User asks a question, and the Assistant solves it. The Assistant first thinks about the reasoning process in the mind and then provides the User with the answer. The reasoning process is enclosed within \textless think\textgreater \textless/think\textgreater and answer is enclosed within \textless answer\textgreater \textless/answer\textgreater tags, respectively, i.e., \textless think\textgreater reasoning process here \textless/think\textgreater \textless answer\textgreater answer here \textless/answer\textgreater. Inside the \textless answer\textgreater...\textless/answer\textgreater block, the final answer must be enclosed in \texttt{\textbackslash boxed\{\}}.\\
\textless\textbar{}User\textbar{}\textgreater\{question\}\\
\textless\textbar{}Assistant\textbar{}\textgreater
}
& DeepSeek-Style &
\begin{lstlisting}[language=Python]
^^Jdef format_reward(completion):^^J \ \ count = 0.0^^J \ \ if completion.count("<think>") == 1:^^J \ \ \ \ count += 0.25^^J \ \ if completion.count("</think>") == 1:^^J \ \ \ \ count += 0.25^^J \ \ if completion.count("<answer>") == 1:^^J \ \ \ \ count += 0.25^^J \ \ if completion.count("</answer>") == 1:^^J \ \ \ \ count += 0.25^^J \ \ return count
\end{lstlisting}
\\\hline
\noindent\parbox[b]{\hsize}{
\textless bos\textgreater{}A conversation between User and Assistant. The User asks a question, and the Assistant solves it. The Assistant first thinks about the reasoning process in the mind and then provides the User with the answer. The reasoning process is enclosed within \textless think\textgreater \textless/think\textgreater and answer is enclosed within \textless answer\textgreater \textless/answer\textgreater tags, respectively, i.e., \textless think\textgreater reasoning process here \textless/think\textgreater \textless answer\textgreater answer here \textless/answer\textgreater. Inside the \textless answer\textgreater...\textless/answer\textgreater block, the final answer must be enclosed in \texttt{\textbackslash boxed\{\}}.\\
\textless\textbar{}User\textbar{}\textgreater\{question\}\\
\textless\textbar{}Assistant\textbar{}\textgreater\textless think\textgreater
}
& DeepSeek-Style (Teacher-Forced) &
\begin{lstlisting}[language=Python]
^^Jdef format_reward(completion):^^J \ \ count = 0.0^^J \ \ if completion.count("</think>") == 1:^^J \ \ \ \ count += 1/3^^J \ \ if completion.count("<answer>") == 1:^^J \ \ \ \ count += 1/3^^J \ \ if completion.count("</answer>") == 1:^^J \ \ \ \ count += 1/3^^J \ \ return count
\end{lstlisting}
\\\hline
\noindent\parbox[b]{\hsize}{
\textless bos\textgreater{}You are an intelligent assistant who helps with user questions. Provide a rigorous, step-by-step derivation of the solution. The final answer must be clearly indicated within \texttt{\textbackslash boxed\{\}}.\\
\textless\textbar{}User\textbar{}\textgreater\{question\}\\
\textless\textbar{}Assistant\textbar{}\textgreater
}
& Freeform &
\begin{lstlisting}[language=Python]
^^Jdef format_reward(completion):^^J \ \ return 1.0
\end{lstlisting}
\\\hline
\noindent\parbox[b]{\hsize}{
\textless bos\textgreater{}Solve the following math challenge. Explain your approach step-by-step\\
The answer should end with: The final answer is: \texttt{\textbackslash boxed\{answer\}}\\
where [answer] is just the final number or expression that solves the problem\\
\textless\textbar{}User\textbar{}\textgreater\{question\}\\
\textless\textbar{}Assistant\textbar{}\textgreater Let's think step by step
}
& Explicit CoT &
\begin{lstlisting}[language=Python]
^^Jdef format_reward(completion):^^J \ \ key = "The final answer is:"^^J \ \ format_score = 1.0 if completion.count(key) == 1 else 0.0^^J \ \ return format_score
\end{lstlisting}
\\\hline
\noindent\parbox[b]{\hsize}{
\textless bos\textgreater{}Analyze and solve the math task.\\
\textless\textbar{}User\textbar{}\textgreater\{question\}\\
End the answer with:\\
The final answer is: \texttt{\textbackslash boxed\{answer\}} where [answer] is just the final number or expression that solves the problem.\\
\textless\textbar{}Assistant\textbar{}\textgreater
}
& Freeform &
\begin{lstlisting}[language=Python]
^^Jdef format_reward(completion):^^J \ \ key = "The final answer is:"^^J \ \ format_score = 1.0 if completion.count(key) == 1 else 0.0^^J \ \ return format_score
\end{lstlisting}
\\\hline
\noindent\parbox[b]{\hsize}{
\textless bos\textgreater{}Solve the following math problem\\
Show each step of your solution\\
Put the final answer within \texttt{\textbackslash boxed\{\}}\\
\textless\textbar{}User\textbar{}\textgreater\{question\}\\
\textless\textbar{}Assistant\textbar{}\textgreater Let's think step by step
}
& Explicit CoT &
\begin{lstlisting}[language=Python]
^^Jdef format_reward(completion):^^J \ \ return 1.0
\end{lstlisting}
    \\\hline
\noindent\parbox[b]{\hsize}{
\textless bos\textgreater
Solve step by step and ensure the final answer is enclosed in \texttt{\textbackslash boxed\{\}}.\\
\textless\textbar{}User\textbar{}\textgreater\{question\}\\
\textless\textbar{}Assistant\textbar{}\textgreater Let's solve this step by step.
}
& Explicit CoT &
\begin{lstlisting}[language=Python]
^^Jdef format_reward(completion):^^J \ \ return 1.0
\end{lstlisting}
\\\hline
\noindent\parbox[b]{\hsize}{
\textless bos\textgreater
Carefully derive the solution step by step. Final answer must be in \texttt{\textbackslash boxed\{\}}.\\
\textless\textbar{}User\textbar{}\textgreater\{question\}\\
\textless\textbar{}Assistant\textbar{}\textgreater
}
& Freeform &
\begin{lstlisting}[language=Python]
^^Jdef format_reward(completion):^^J \ \ return 1.0
\end{lstlisting}
\\\hline
\noindent\parbox[b]{\hsize}{
\textless bos\textgreater
Provide a detailed reasoning process. Conclude with \texttt{\textbackslash boxed\{\}}.\\
\textless\textbar{}User\textbar{}\textgreater\{question\}\\
\textless\textbar{}Assistant\textbar{}\textgreater Start reasoning:
}
& Explicit CoT &
\begin{lstlisting}[language=Python]
^^Jdef format_reward(completion):^^J \ \ return 1.0
\end{lstlisting}
\\\hline
\noindent\parbox[b]{\hsize}{
\textless bos\textgreater
Break the problem into steps and solve sequentially. Final answer in \texttt{\textbackslash boxed\{\}}.\\
\textless\textbar{}User\textbar{}\textgreater\{question\}\\
\textless\textbar{}Assistant\textbar{}\textgreater
}
& Freeform &
\begin{lstlisting}[language=Python]
^^Jdef format_reward(completion):^^J \ \ return 1.0
\end{lstlisting}
\\\hline
\noindent\parbox[b]{\hsize}{
\textless bos\textgreater
Follow strict reasoning. Do not skip steps. Final answer must be in \texttt{\textbackslash boxed\{\}}.\\
\textless\textbar{}User\textbar{}\textgreater\{question\}\\
\textless\textbar{}Assistant\textbar{}\textgreater
}
& Freeform &
\begin{lstlisting}[language=Python]
^^Jdef format_reward(completion):^^J \ \ return 1.0
\end{lstlisting}
\\\hline
\noindent\parbox[b]{\hsize}{
\textless bos\textgreater
Explain clearly and end precisely with \texttt{\textbackslash boxed\{\}}.\\
\textless\textbar{}User\textbar{}\textgreater\{question\}\\
\textless\textbar{}Assistant\textbar{}\textgreater Proceed carefully:
}
& Explicit CoT &
\begin{lstlisting}[language=Python]
^^Jdef format_reward(completion):^^J \ \ return 1.0
\end{lstlisting}
\\\hline
\noindent\parbox[b]{\hsize}{
\textless bos\textgreater
Ensure logical correctness at every step. Final answer strictly inside \texttt{\textbackslash boxed\{\}}.\\
\textless\textbar{}User\textbar{}\textgreater\{question\}\\
\textless\textbar{}Assistant\textbar{}\textgreater
}
& Freeform &
\begin{lstlisting}[language=Python]
^^Jdef format_reward(completion):^^J \ \ return 1.0
\end{lstlisting}
\\\hline
\noindent\parbox[b]{\hsize}{
\textless bos\textgreater
Solve the problem and show reasoning. End with:\\
We conclude that the answer is: \texttt{\textbackslash boxed\{answer\}}\\
\textless\textbar{}User\textbar{}\textgreater\{question\}\\
\textless\textbar{}Assistant\textbar{}\textgreater
}
& Freeform &
\begin{lstlisting}[language=Python]
^^Jdef format_reward(completion):^^J \ \ key = "We conclude that the answer is:"^^J \ \ return 1.0 if completion.count(key) == 1 else 0.0
\end{lstlisting}
\\\hline
\noindent\parbox[b]{\hsize}{
\textless bos\textgreater
Work through the math problem step by step. Finish with:\\
Therefore, the final answer is \texttt{\textbackslash boxed\{answer\}}\\
\textless\textbar{}User\textbar{}\textgreater\{question\}\\
\textless\textbar{}Assistant\textbar{}\textgreater Let's think carefully.
}
& Explicit CoT &
\begin{lstlisting}[language=Python]
^^Jdef format_reward(completion):^^J \ \ key = "Therefore, the final answer is"^^J \ \ return 1.0 if completion.count(key) == 1 else 0.0
\end{lstlisting}
\\\hline
\noindent\parbox[b]{\hsize}{
\textless bos\textgreater
Provide reasoning and ensure the answer ends with:\\
Based on the analysis, the answer is: \texttt{\textbackslash boxed\{answer\}}\\
\textless\textbar{}User\textbar{}\textgreater\{question\}\\
\textless\textbar{}Assistant\textbar{}\textgreater
}
& Freeform &
\begin{lstlisting}[language=Python]
^^Jdef format_reward(completion):^^J \ \ key = "Based on the analysis, the answer is:"^^J \ \ return 1.0 if completion.count(key) == 1 else 0.0
\end{lstlisting}
\\\hline
\noindent\parbox[b]{\hsize}{
\textless bos\textgreater
Explain your reasoning step by step, and conclude with a final statement of the form:\\
From the steps above, the answer is \texttt{\textbackslash boxed\{answer\}}\\
\textless\textbar{}User\textbar{}\textgreater\{question\}\\
\textless\textbar{}Assistant\textbar{}\textgreater
}
& Freeform &
\begin{lstlisting}[language=Python]
^^Jdef format_reward(completion):^^J \ \ key = "From the steps above, the answer is"^^J \ \ return 1.0 if completion.count(key) == 1 else 0.0
\end{lstlisting}
\\\hline
\noindent\parbox[b]{\hsize}{
\textless bos\textgreater
Provide concise but sufficient reasoning. Final answer must be in \texttt{\textbackslash boxed\{\}}.\\
\textless\textbar{}User\textbar{}\textgreater\{question\}\\
\textless\textbar{}Assistant\textbar{}\textgreater
}
& (Concise) Freeform &
\begin{lstlisting}[language=Python]
^^Jdef format_reward(completion):^^J \ \ return 1.0
\end{lstlisting}
\\\hline
\noindent\parbox[b]{\hsize}{
\textless bos\textgreater
Solve briefly but include key steps. End with \texttt{\textbackslash boxed\{\}}.\\
\textless\textbar{}User\textbar{}\textgreater\{question\}\\
\textless\textbar{}Assistant\textbar{}\textgreater
}
& (Concise) Freeform &
\begin{lstlisting}[language=Python]
^^Jdef format_reward(completion):^^J \ \ return 1.0
\end{lstlisting}
\\\hline
\noindent\parbox[b]{\hsize}{
\textless bos\textgreater
Solve the following problem.\\
Format:\\
\textless think\textgreater\\
...\\
\textless/think\textgreater\\
\textless answer\textgreater\\
\texttt{\textbackslash boxed\{\}}\\
\textless/answer\textgreater\\
Ensure correctness and proper formatting.\\
\textless\textbar{}User\textbar{}\textgreater\{question\}\\
\textless\textbar{}Assistant\textbar{}\textgreater\textless think\textgreater
}
& DeepSeek-Style (Teacher-Forced) &
\begin{lstlisting}[language=Python]
^^Jdef format_reward(completion):^^J \ \ count = 0.0^^J \ \ if completion.count("\n</think>\n") == 1:^^J \ \ \ \ count += 1/3^^J \ \ if completion.count("\n<answer>\n") == 1:^^J \ \ \ \ count += 1/3^^J \ \ if completion.count("\n</answer>") == 1:^^J \ \ \ \ count += 1/3^^J \ \ return count
\end{lstlisting}
\\\hline
\noindent\parbox[b]{\hsize}{
\textless bos\textgreater
Solve the following problem.\\
Format:\\
\textless think\textgreater\\
...\\
\textless/think\textgreater\\
\textless answer\textgreater\\
\texttt{\textbackslash boxed\{\}}\\
\textless/answer\textgreater\\
Ensure correctness and proper formatting.\\
\textless\textbar{}User\textbar{}\textgreater\{question\}\\
\textless\textbar{}Assistant\textbar{}\textgreater
}
& DeepSeek-Style &
\begin{lstlisting}[language=Python]
^^Jdef format_reward(completion):^^J \ \ count = 0.0^^J \ \ if completion.count("<think>\n") == 1:^^J \ \ \ \ count += 0.25^^J \ \ if completion.count("\n</think>\n") == 1:^^J \ \ \ \ count += 0.25^^J \ \ if completion.count("\n<answer>\n") == 1:^^J \ \ \ \ count += 0.25^^J \ \ if completion.count("\n</answer>") == 1:^^J \ \ \ \ count += 0.25^^J \ \ return count
\end{lstlisting}
\\\hline
\noindent\parbox[b]{\hsize}{
\textless bos\textgreater
Solve the problem carefully and ensure correctness.\\
Use this exact format:\\
\textless think\textgreater...\textless/think\textgreater\textless answer\textgreater...\textless/answer\textgreater\\
The answer must be enclosed in \texttt{\textbackslash boxed\{\}}.\\
All tags must be properly closed.\\
\textless\textbar{}User\textbar{}\textgreater\{question\}\\
\textless\textbar{}Assistant\textbar{}\textgreater\textless think\textgreater
}
& DeepSeek-Style (Teacher-Forced) &
\begin{lstlisting}[language=Python]
^^Jdef format_reward(completion):^^J \ \ count = 0.0^^J \ \ if completion.count("</think>") == 1:^^J \ \ \ \ count += 1/3^^J \ \ if completion.count("<answer>") == 1:^^J \ \ \ \ count += 1/3^^J \ \ if completion.count("</answer>") == 1:^^J \ \ \ \ count += 1/3^^J \ \ return count
\end{lstlisting}
\\\hline
\noindent\parbox[b]{\hsize}{
\textless bos\textgreater
Solve the problem carefully and ensure correctness.\\
Use this exact format:\\
\textless think\textgreater...\textless/think\textgreater\textless answer\textgreater...\textless/answer\textgreater\\
The answer must be enclosed in \texttt{\textbackslash boxed\{\}}.\\
All tags must be properly closed.\\
\textless\textbar{}User\textbar{}\textgreater\{question\}\\
\textless\textbar{}Assistant\textbar{}\textgreater
}
& DeepSeek-Style &
\begin{lstlisting}[language=Python]
^^Jdef format_reward(completion):^^J \ \ count = 0.0^^J \ \ if completion.count("<think>") == 1:^^J \ \ \ \ count += 0.25^^J \ \ if completion.count("</think>") == 1:^^J \ \ \ \ count += 0.25^^J \ \ if completion.count("<answer>") == 1:^^J \ \ \ \ count += 0.25^^J \ \ if completion.count("</answer>") == 1:^^J \ \ \ \ count += 0.25^^J \ \ return count
\end{lstlisting}
\\\hline
\noindent\parbox[b]{\hsize}{
\textless bos\textgreater
You are given a problem to solve.\\
Use this structure exactly:\\
\textless think\textgreater\\
...\\
\textless/think\textgreater\\
\textless answer\textgreater\\
...\\
\textless/answer\textgreater\\
Final answer must be enclosed in \texttt{\textbackslash boxed\{\}}.\\
Do not skip steps.\\
\textless\textbar{}User\textbar{}\textgreater\{question\}\\
\textless\textbar{}Assistant\textbar{}\textgreater\textless think\textgreater
}
& DeepSeek-Style (Teacher-Forced) &
\begin{lstlisting}[language=Python]
^^Jdef format_reward(completion):^^J \ \ count = 0.0^^J \ \ if completion.count("</think>") == 1:^^J \ \ \ \ count += 1/3^^J \ \ if completion.count("<answer>") == 1:^^J \ \ \ \ count += 1/3^^J \ \ if completion.count("</answer>") == 1:^^J \ \ \ \ count += 1/3^^J \ \ return count
\end{lstlisting}
\\\hline
\noindent\parbox[b]{\hsize}{
\textless bos\textgreater
You are given a problem to solve.\\
Use this structure exactly:\\
\textless think\textgreater\\
...\\
\textless/think\textgreater\\
\textless answer\textgreater\\
...\\
\textless/answer\textgreater\\
Final answer must be enclosed in \texttt{\textbackslash boxed\{\}}.\\
Do not skip steps.\\
\textless\textbar{}User\textbar{}\textgreater\{question\}\\
\textless\textbar{}Assistant\textbar{}\textgreater
}
& DeepSeek-Style &
\begin{lstlisting}[language=Python]
^^Jdef format_reward(completion):^^J \ \ count = 0.0^^J \ \ if completion.count("<think>") == 1:^^J \ \ \ \ count += 0.25^^J \ \ if completion.count("</think>") == 1:^^J \ \ \ \ count += 0.25^^J \ \ if completion.count("<answer>") == 1:^^J \ \ \ \ count += 0.25^^J \ \ if completion.count("</answer>") == 1:^^J \ \ \ \ count += 0.25^^J \ \ return count
\end{lstlisting}
\\\hline
\noindent\parbox[b]{\hsize}{
\textless bos\textgreater
Your task is to solve the given problem correctly.\\
All reasoning must be inside \textless think\textgreater\textless/think\textgreater.\\
The final answer must be enclosed in \texttt{\textbackslash boxed\{\}}.\\
Ensure correctness before answering.\\
\textless\textbar{}User\textbar{}\textgreater\{question\}\\
\textless\textbar{}Assistant\textbar{}\textgreater\textless think\textgreater
}
& DeepSeek-Style (Teacher-Forced) &
\begin{lstlisting}[language=Python]
^^Jdef format_reward(completion):^^J \ \ count = 0.0^^J \ \ if completion.count("</think>") == 1:^^J \ \ \ \ count += 1^^J \ \ return count
\end{lstlisting}
\\\hline
\noindent\parbox[b]{\hsize}{
\textless bos\textgreater
Your task is to solve the given problem correctly.\\
All reasoning must be inside \textless think\textgreater\textless/think\textgreater.\\
The final answer must be enclosed in \texttt{\textbackslash boxed\{\}}.\\
Ensure correctness before answering.\\
\textless\textbar{}User\textbar{}\textgreater\{question\}\\
\textless\textbar{}Assistant\textbar{}\textgreater
}
& DeepSeek-Style &
\begin{lstlisting}[language=Python]
^^Jdef format_reward(completion):^^J \ \ count = 0.0^^J \ \ if completion.count("<think>") == 1:^^J \ \ \ \ count += 0.5^^J \ \ if completion.count("</think>") == 1:^^J \ \ \ \ count += 0.5^^J \ \ return count
\end{lstlisting}
\\\hline
\noindent\parbox[b]{\hsize}{
\textless bos\textgreater
Solve the problem step by step.\\
Organize your reasoning clearly inside \textless think\textgreater\textless/think\textgreater.\\
The final answer must be in \texttt{\textbackslash boxed\{\}}.\\
\textless\textbar{}User\textbar{}\textgreater\{question\}\\
\textless\textbar{}Assistant\textbar{}\textgreater\textless think\textgreater
}
& DeepSeek-Style (Teacher-Forced) &
\begin{lstlisting}[language=Python]
^^Jdef format_reward(completion):^^J \ \ count = 0.0^^J \ \ if completion.count("</think>") == 1:^^J \ \ \ \ count += 1^^J \ \ return count
\end{lstlisting}
\\\hline
\noindent\parbox[b]{\hsize}{
\textless bos\textgreater
Solve the problem step by step.\\
Organize your reasoning clearly inside \textless think\textgreater\textless/think\textgreater.\\
The final answer must be in \texttt{\textbackslash boxed\{\}}.\\
\textless\textbar{}User\textbar{}\textgreater\{question\}\\
\textless\textbar{}Assistant\textbar{}\textgreater
}
& DeepSeek-Style &
\begin{lstlisting}[language=Python]
^^Jdef format_reward(completion):^^J \ \ count = 0.0^^J \ \ if completion.count("<think>") == 1:^^J \ \ \ \ count += 0.5^^J \ \ if completion.count("</think>") == 1:^^J \ \ \ \ count += 0.5^^J \ \ return count
\end{lstlisting}
\\\hline
\noindent\parbox[b]{\hsize}{
\textless bos\textgreater{}Please reason step by step, and put your final answer within \texttt{\textbackslash boxed\{\}}.\\
\textless\textbar{}User\textbar{}\textgreater\{question\}\\
\textless\textbar{}Assistant\textbar{}\textgreater\textless think\textgreater
}
& DeepSeek-style (Default Template) &
\begin{lstlisting}[language=Python]
^^Jdef format_reward(completion):^^J \ \ count = 0.0^^J \ \ if completion.count("</think>") == 1:^^J \ \ \ \ count += 1^^J \ \ return count
\end{lstlisting}
\\\hline
\end{tabularx}

\newpage
\subsection{Format Reward Best Practices}
\label{sec:format_prac}
Empirically, we find that simple additive format rewards often achieve the best performance. In contrast, more complex conditional logic or regular-expression-based rewards can degrade reasoning accuracy. In particular, rules that overly restrict the model’s behavior may accelerate training collapse and lead to worse outcomes. For example, the format reward in \cref{lst:bad_reward} limits the number of characters generated after the model’s final boxed answer to fewer than 70, with the goal of preventing the model from continuing to output gibberish after completing its answer. However, we find that this reward structure can induce early collapse and substantially degrade performance.

\begin{lstlisting}[language=Python, caption={abel\_format\_reward is an example of an overly complex and restrictive format reward function, which requires the model to generate fewer than 70 characters after its final answer.}, label={lst:bad_reward}]
def find_matching_brace(s, start):
    depth = 1   # assumes s[start] == "{"
    for i in range(start + 1, len(s)):
        c = s[i]
        if c == "{":
            depth += 1
        elif c == "}":
            depth -= 1
            if depth == 0:
                return i
    return -1

def boxed_format_and_length_penalty(completion):
    key = r"\boxed"
    idx = completion.rfind(key)
    if idx == -1:
        return 0.0

    # find opening brace after \boxed
    brace_start = completion.find("{", idx)
    if brace_start == -1:
        return 0.0

    # find matching closing brace
    brace_end = find_matching_brace(completion, brace_start)
    if brace_end == -1:
        return 0.0

    # count chars after boxed expression
    tail_len = len(completion) - (brace_end + 1)

    if tail_len > 70:
        return 0.0

    return 1.0

# should be very limited words after last boxed
def abel_format_reward(completion):
    return boxed_format_and_length_penalty(completion)
\end{lstlisting}

\end{document}